\documentclass[letterpaper]{article} 
\usepackage{aaai24}  
\usepackage{times}  
\usepackage{helvet}  
\usepackage{courier}  
\usepackage[hyphens]{url}  
\usepackage{graphicx} 
\usepackage{natbib}  
\usepackage{caption} 
\usepackage{algorithm}
\usepackage{algorithmic}

\usepackage{amsmath}
\usepackage{booktabs}
\usepackage{multirow}
\usepackage{amssymb}

\usepackage{newfloat}
\usepackage{listings}
\usepackage{makecell}
\DeclareCaptionStyle{ruled}
{labelfont=normalfont,labelsep=colohttps://www.overleaf.com/project/64af749dfad7c734cc92de76n,strut=off} 
\floatstyle{ruled}
\newfloat{listing}{tb}{lst}{}
\floatname{listing}{Listing}
\title{Text Image Inpainting via Global Structure-Guided Diffusion Models}

\author{
    Shipeng Zhu\textsuperscript{\rm 1,2}, Pengfei Fang\textsuperscript{\rm 1,2}, Chenjie Zhu\textsuperscript{\rm 1,2}, Zuoyan Zhao\textsuperscript{\rm 1,2}, Qiang Xu\textsuperscript{\rm 1,2}, Hui Xue\textsuperscript{\rm 1,2}\thanks{Corresponding author.}
}

\affiliations{
    \textsuperscript{\rm 1}School of Computer Science and Engineering, Southeast University, Nanjing 210096, China\\
    \textsuperscript{\rm 2}Key Laboratory of New Generation Artificial Intelligence Technology and Its Interdisciplinary\\ Applications (Southeast University), Ministry of Education, China\\
    
    \{shipengzhu, fangpengfei, chenjiezhu, zuoyanzhao, 220232307, hxue\}@seu.edu.cn

}

\usepackage{bibentry}

\begin{document}

\maketitle

\begin{abstract}
Real-world text can be damaged by corrosion issues caused by environmental or human factors, which hinder the preservation of the complete styles of texts, e.g., texture and structure. These corrosion issues, such as graffiti signs and incomplete signatures, bring difficulties in understanding the texts, thereby posing significant challenges to downstream applications, e.g., scene text recognition and signature identification. Notably, current inpainting techniques often fail to adequately address this problem and have difficulties restoring accurate text images along with reasonable and consistent styles. Formulating this as an open problem of text image inpainting, this paper aims to build a benchmark to facilitate its study. In doing so, we establish two specific text inpainting datasets which contain scene text images and handwritten text images, respectively. Each of them includes images revamped by real-life and synthetic datasets, featuring pairs of original images, corrupted images, and other assistant information. On top of the datasets, we further develop a novel neural framework, Global Structure-guided Diffusion Model (GSDM), as a potential solution. Leveraging the global structure of the text as a prior, the proposed GSDM develops an efficient diffusion model to recover clean texts. The efficacy of our approach is demonstrated by thorough empirical study, including a substantial boost in both recognition accuracy and image quality. These findings not only highlight the effectiveness of our method but also underscore its potential to enhance the broader field of text image understanding and processing. Code and datasets are available at: https://github.com/blackprotoss/GSDM.
\end{abstract}

\section{Introduction}

\begin{figure}[t]
\centering
\includegraphics[width=1\columnwidth]{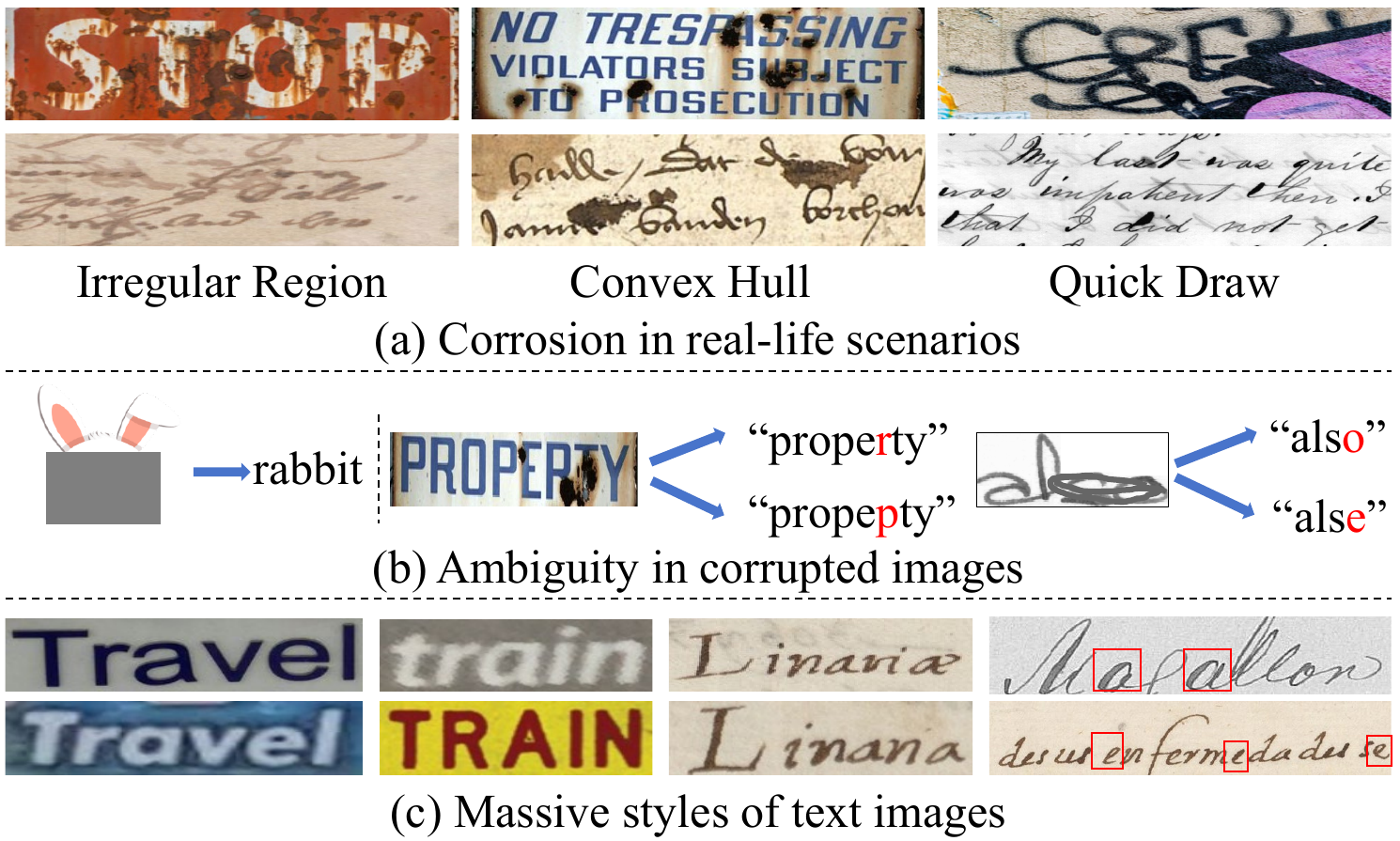} 
\caption{The illustration of corrosion forms in real-life scenarios and the challenges of text image inpainting.}
\label{fig:realdata}
\end{figure}

Text in the real world serves as a visual embodiment of human language~\cite{long2021scene}. It plays a vital role in conveying vast linguistic information and facilitating communication and collaboration in daily life. However, the integrity of text with specific styles, e.g., structure, texture, and background clutter, can be compromised by factors such as environmental corrosion and human interference~\cite{krishnan2023textstylebrush}. As a consequence, these resultant images, as shown in Figure~\ref{fig:realdata}(a), are inherently degraded, leading to a performance drop in the text reading and understanding systems. In other words, tasks such as scene text editing~\cite{qu2023exploring} and signature verification~\cite{lai2021synsig2vec} are inevitably affected by the integrity of text images.

Aiming to provide visually plausible restoration for missing regions in corrupted images~\cite{bertalmio2003simultaneous,xiang2023deep}, image inpainting technologies have made considerable progress~\cite{zhao2022transcnn,lugmayr2022repaint,ji2023improving,yu2023inpaint} in recent years. However, some inherent challenges restrict these general image inpainting methods from restoring corrupted text images.  Firstly, \textbf{the corrupted regions of text images are unknown.} That is, the corrosive factors, rooted in real-life scenarios, mean the location mask cannot be provided. Consequently, prevailing non-blind inpainting methods cannot handle this entire image reconstruction task. Secondly, \textbf{the corrupted regions induce content ambiguity in the text image.} It is known that natural objects can be recognized based on their iconic local features. For example, a rabbit can be easily recognized by its long ears, despite corrosion over most of the body parts (Shown in Figure~\ref{fig:realdata}(b)). However, the corrosion disrupts the integrity of the global structure in the text image, including its shape and profile, making it challenging to reconstruct the correct characters/words from the remaining strokes. Lastly, \textbf{text images contain massive style variations}. The text images exhibit high inter- and intra-class variability in style~\cite{krishnan2023textstylebrush}, with variations spanning background properties, typography, etc. For instance, two characters of the same class may appear differently, even within the same image (See the red rectangles in Figure~\ref{fig:realdata}(c)). This reality places substantial demands on the generalization of a machine to repair corrupted text images.

\begin{figure}[t]
\centering
\includegraphics[width=1\columnwidth]{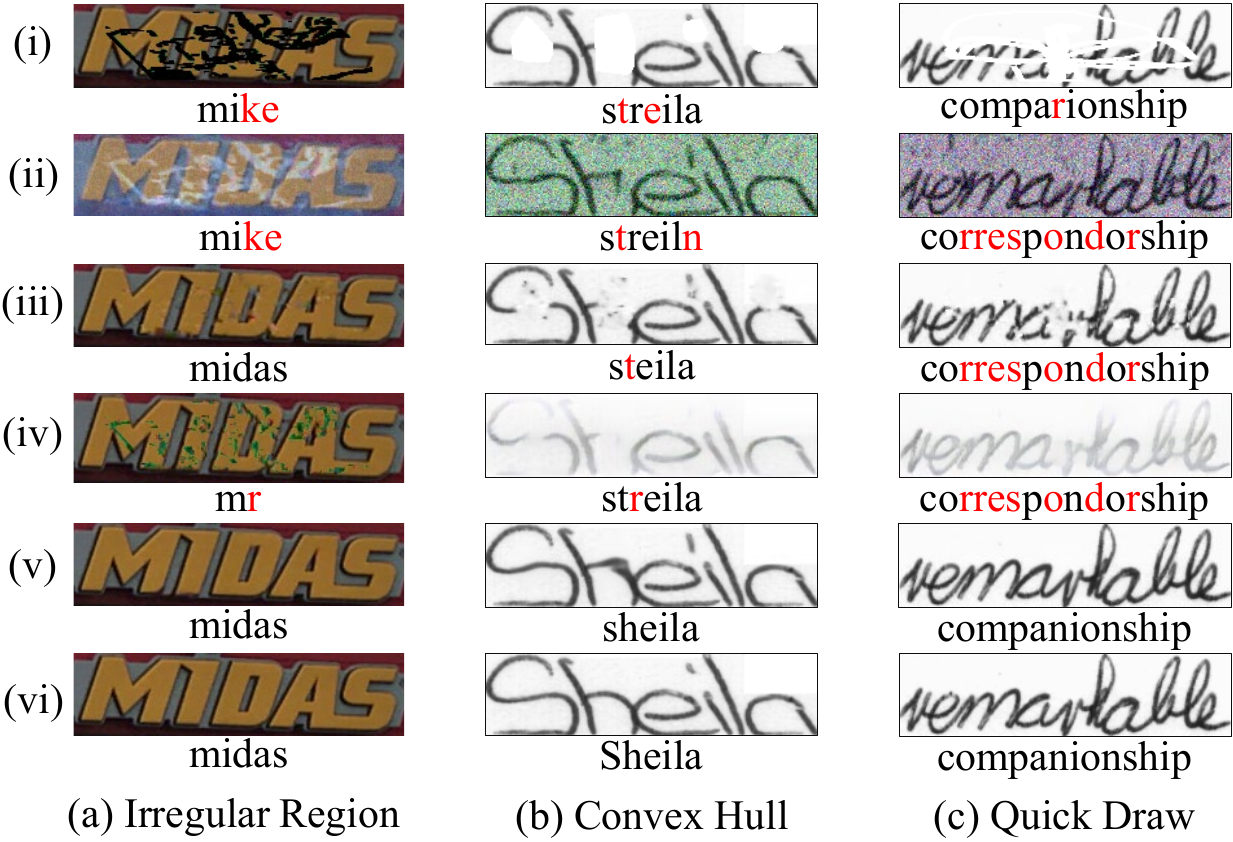} 
\caption{The illustration of inpainting images with recognition results based on different methods. The (i) to (vi) denote Corrupted images, DDIM, CoPaint, TransCNN-HAE. GSDM, and GT. Red characters indicate errors.}
\label{fig:inpaintingexample}
\end{figure}

This paper investigates this challenging task, named \textbf{text image inpainting}, and addresses it by formally formulating the task and establishing a benchmark. The closest study to our work is~\cite{sun2022tsinit}, which introduces a scene text image dataset for foreground text completion. However, it only includes one corrosion form for synthetic images, thus failing to reflect diverse real-world scenes effectively. \textbf{Realizing the gaps, our study takes a deep dive, with a focus on restoring the real corrupted text images.} As a result, one can enable the restoration of style and detail consistency in corrupted text images, as illustrated in Figure~\ref{fig:inpaintingexample}. Aligning with the paradigm used in tailored text image tasks~\cite{wu2019editing}, we gather real-life and synthetic text images to produce two tailored datasets: the Scene Text Image Inpainting (TII-ST) dataset and Handwritten Text Image Inpainting (TII-HT) dataset. In these datasets, we design three typical corrosion forms, i.e., convex hull, irregular region, and quick draw, affecting both scene text images and handwritten text images. With these enriched datasets, we can evaluate the image quality produced by various inpainting methods and assess their impact on downstream applications.

Along with the datasets, we further propose a simple yet effective neural network, dubbed Global Structure-guided Diffusion Model (GSDM), as a baseline for the text image inpainting task. The proposed GSDM leverages the structure of the text as a prior, guiding the diffusion model in realizing image restoration. To this end, a Structure Prediction Module (SPM) is first proposed to generate a complete segmentation map that offers guidance regarding the content and positioning of the text. The subsequent diffusion-based Reconstruction Module (RM), which receives the predicted segmentation mask and corrupted images as input, is developed to generate intact text images with coherent styles efficiently. As shown in Figure~\ref{fig:inpaintingexample}, our proposed GSDM outperforms comparison methods and generates plausible images.
In a nutshell, our \textbf{contributions} are as follows: 

\begin{itemize}
\item We construct two datasets, TII-ST and TII-HT, which facilitate the study of text image inpainting. To our knowledge, this is the first initiative to fully restore all styles of corrupted text images, thereby defining a challenging yet promising task.

\item We propose a versatile method, the Global Structure-guided Diffusion Model (GSDM), as a baseline for the task. This model uses the guidance of the complete global structure, predicted from the remaining regions of corrupted text images, to generate complete text images coherent with the corrupted ones.

\item Comparisons with relevant approaches on the TII-ST and TII-HT datasets demonstrate that our GSDM outperforms these approaches in enhancing downstream applications and improving image quality. Substantial ablation studies further underscore the necessity of different components in our model. The realistic benchmark and strong performance of our work provide favorable templates for future research. 

\end{itemize}

\section{Related Work}

\begin{table*}[t]
\centering
\resizebox{1\textwidth}{!}{
\begin{tabular}{c|c|c|c|c|c}
\Xhline{1.2pt}
Dataset & Data Type & Image Number & Corrosion Form & Corrosion Ratio Range & Evaluation Protocal \\ \cline{1-6} 
TII-ST & Synthesis Image + Real Image & 86,476 & CH + IR + QD & 5\%--60\% &  Accuracy + Quality \\
TII-HT & Real Image & 40,078 & CH + IR + QD & 5\%--60\% & Accuracy + Quality \\ 
\Xhline{1.2pt}
\end{tabular}}
\caption{The data statistics of two constructed datasets, TII-ST and TII-HT. The ``CH", ``IR", and ``QD" denote convex hull, irregular region, and quick draw, respectively. The ``Accuracy" denotes the word-level recognition accuracy.}
  \label{tab:analysisofdata}
\end{table*}

\subsection{Image Inpainting}

Image inpainting has long posed a challenge within the computer vision community, aiming for the coherent restoration of corrupted images~\cite{shah2022overview,xiang2023deep}. In earlier developments, the majority of approaches have grounded their foundations in auto-encoders~\cite{yu2022unbiased}, auto-regressive transformers~\cite{wan2021high}, and GAN-based paradigms~\cite{pan2021exploiting}. Notably, diffusion-based techniques~\cite{lugmayr2022repaint,zhang2023towards,yu2023inpaint} have recently gained attention due to their exceptional capability in image generation~\cite{ramesh2022hierarchical}. Within this context, CoPaint~\cite{zhang2023towards} presents a Bayesian framework for holistic image modification, achieving state-of-the-art performance in natural image inpainting. Yet, these methods necessitate explicit guidance of the corrupted mask, which hinders their adaptability in real-world contexts. Moreover, there have been endeavors centered on blind inpainting, which eschew reliance on provided corrupted masks, addressing challenges through image-to-image paradigms~\cite{cai2017blind,zhang2017demeshnet,wang2020vcnet}. For instance, TransCNN-HAE~\cite{zhao2022transcnn} innovatively employs a hybrid Transformer-CNN auto-encoder, optimizing the capability to excavate both long and short range contexts. 
Concurrently, some diffusion-oriented models~\cite{kawar2022denoising,fei2023generative} with a dedication to unified image restoration have showcased capabilities in blind image inpainting. However, all these methods are primarily suitable for natural images, thus making it difficult to handle text images, whose semantics are sensitive to the text structure.

Zooming into tailored character inpainting, notable progress~\cite{chang2018chinese} has been made. Recently, Wang et al. leverage the semantic acuity of BERT~\cite{devlin2018bert}, reconstructing the corrupted strokes inherent in Chinese characters~\cite{wang2021image}. Moreover, TSINIT~\cite{sun2022tsinit} proposes a two-stage encoder-decoder blueprint, generating intact binary foreground texts from incomplete scene text images. Nonetheless, it is worth noting that such methods merely focus on the structure of text images. They overlook the diverse styles inherent in text images, 
which impacts human perception and narrows downstream applications.

\subsection{Text Image Recognition}

Text image recognition serves as a foundational element for complicated text-understanding tasks~\cite{he2023icl} and the assessment of image processing endeavors~\cite{wang2020scene,wu2019editing}. Wherein, Scene Text Recognition (STR) and Handwritten Text Recognition (HTR) emerge as dominant research areas~\cite{zhu2023conditional}. Scene text images showcase a myriad of text styles, both in texture and layout. Pioneering in this field, CRNN~\cite{shi2016end} leverages sequential information in scene text images, achieving proficient recognition of variable-length images. Successor models like ASTER~\cite{shi2018aster} and MORAN~\cite{luo2019moran} further enhance recognition performance through diverse visual rectification techniques. More recently, language-aware approaches~\cite{fang2021read,bautista2022parseq} harness the predictive capabilities of language models~\cite{devlin2018bert,yang2019xlnet} to map word probabilities, resulting in impressive recognition outcomes. 

For handwritten text images, they exhibit diverse calligraphic styles, such as joined-up and illegible handwriting. 
In recent advancements, numerous methods~\cite{wang2020decoupled,singh2021full,li2023trocr} tap into attention mechanisms to perceive structural correlations, thereby attaining promising performance.

\section{Benchmark Dataset}

\begin{figure}[t]
\centering
\includegraphics[width=1\columnwidth]{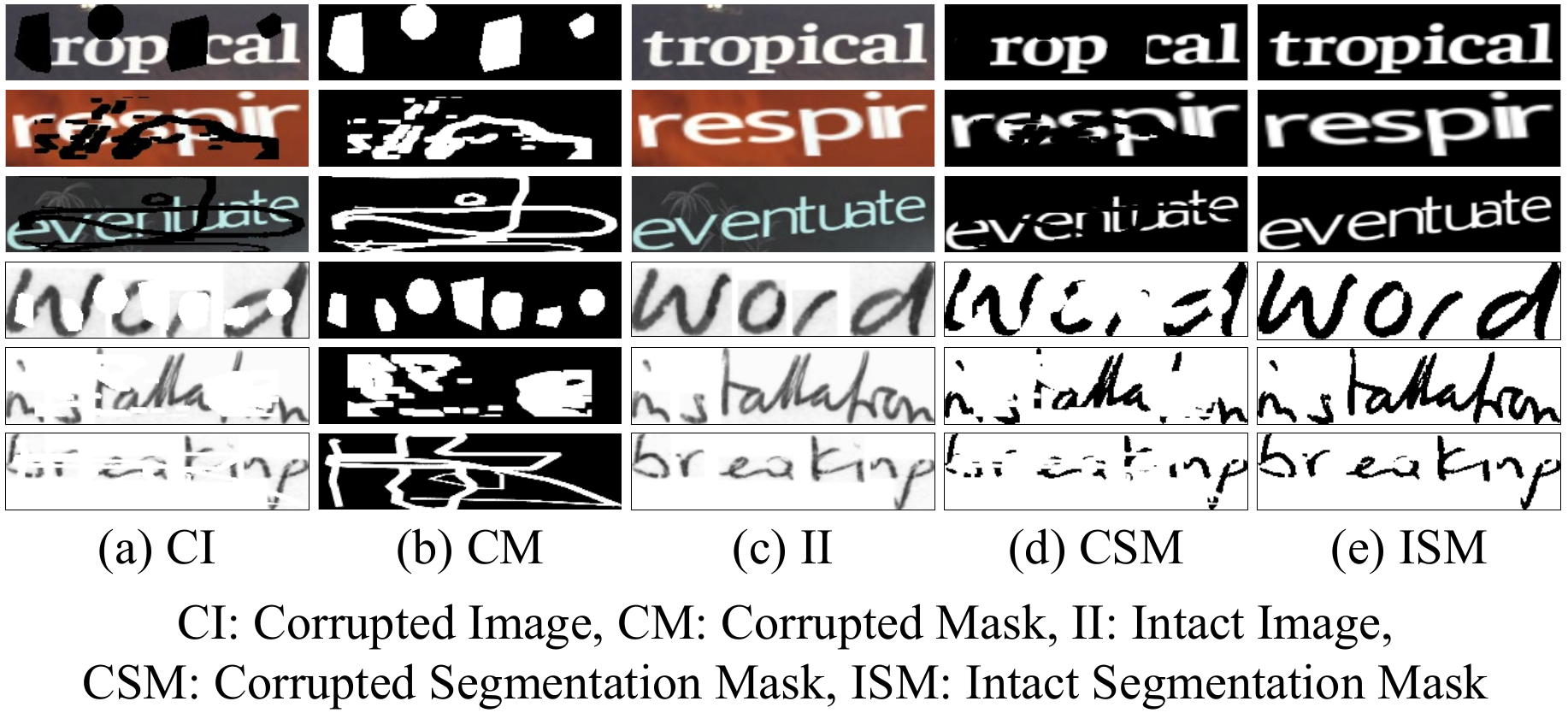} 
\caption{Some training examples in the two datasets. The images of the first three rows are from TII-ST and the images of the last three rows are from TII-HT.} 
\label{fig:dataset}
\end{figure}

\subsection{Dataset Description}

\begin{figure*}[t]
\centering
\includegraphics[width=0.95\textwidth]{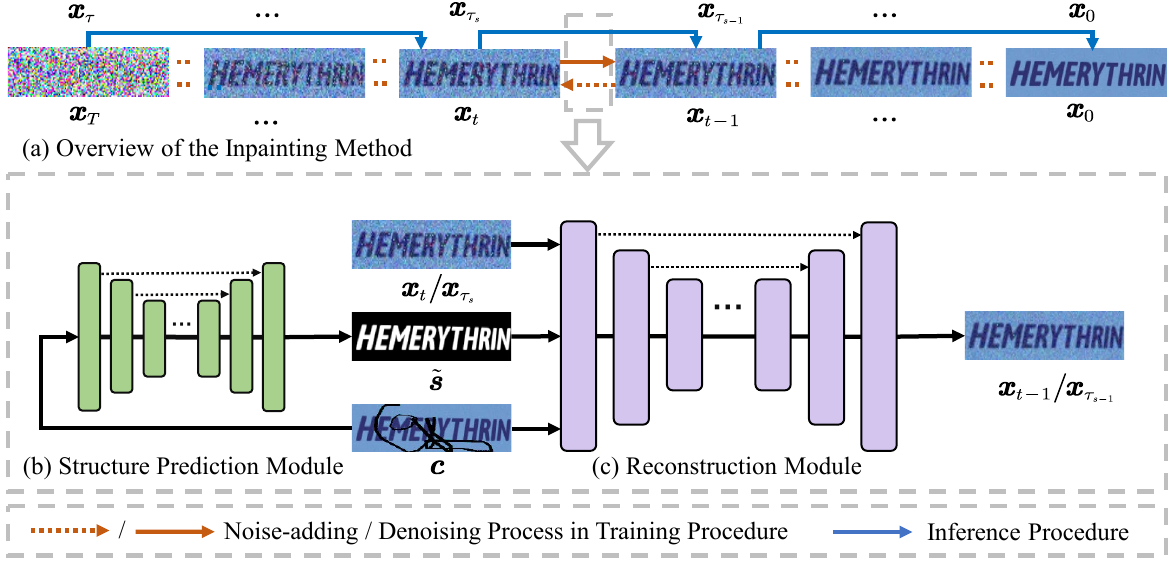} 
\caption{The overall architecture of our proposed Global Structure-guided Diffusion Model (GSDM). It consists of two main modules:  Structure Prediction Module (SPM) and Reconstruction Module (RM). 
} 
\label{fig:archi}
\end{figure*}

Text image inpainting focuses on reconstructing corrupted images, which have been subjected to a variety of real-world disturbances and lack corresponding pristine versions. In this paper, we introduce two novel datasets, TII-ST and TII-HT, tailored for this task. Given the vast style variation in scene text images~\cite{krishnan2023textstylebrush}, we construct the TII-ST dataset using a combination of synthesized and real images. First, we choose to create our own synthetic images instead of utilizing an existing synthetic dataset~\cite{gupta2016synthetic}, to provide rich auxiliary information, of which segmentation masks are introduced to our basic TII-ST. Specifically,  following the method in~\cite{jaderberg2014synthetic}, we synthesize 80,000 scene text images. Next, we supplement the scene text image dataset with 6,476 real scene text images collected from various sources, including ICDAR 2013~\cite{karatzas2013icdar}, ICDAR 2015~\cite{karatzas2015icdar}, and ICDAR 2017~\cite{nayef2017icdar2017}. For handwritten text, the TII-HT dataset comprises 40,078 images from the IAM dataset~\cite{marti2002iam}. The text segmentation mask for each image can be acquired using a predetermined threshold.

To accurately simulate real-life corrosion (See an illustration in Figure~\ref{fig:realdata}), we introduce distinct corrosion forms, i.e., convex hull, irregular region, and quick draw. Notably, the shape of each form can be governed by specific parameters. By adopting these flexible corrosion forms, we aim to encompass a broad spectrum of potential real-world image corrosion scenarios, thereby bolstering the versatility and robustness of the text image inpainting task. Utilizing the images and corrosion forms, we create tuples for each pristine image in both datasets. In the training set, each tuple contains a corrupted image, its corrupted mask, the original intact image, a corrupted segmentation mask, and an intact segmentation mask. For the testing dataset, we furnish data pairs, comprising only the corrupted and intact images. All these images are resized to $64\times256$ to ensure consistent evaluation. Sample images from both datasets are depicted in Figure~\ref{fig:dataset}. Additionally, Table~\ref{tab:analysisofdata} intuitively presents basic statistics of the proposed datasets.

\subsection{Evaluation Protocal}

For fairness in evaluation, we divide our proposed datasets into distinct training and testing sets, respectively.  In the TII-ST dataset, we follow the strategy outlined in~\cite{zhu2023conditional}. Specifically, our training set consists of 80,000 synthesized images and 4,877 real images. Meanwhile, the testing set includes 1,599 real images. For the TII-HT dataset, the training set comprises of 38,578 images sourced from IAM, while the testing set contains 1,600 images.

The evaluation of inpainting results on these datasets takes into account both the impact on downstream tasks and the overall image quality. 
We use text recognition to assess improvements to downstream tasks and employ two established metrics, Peak Signal-to-Noise Ratio (PSNR) (dB) and Structural SIMilarity (SSIM), to evaluate image quality.

Recognizing the profound influence of text image quality on reading and understanding systems~\cite{wang2020scene}, we opt for text recognition as a representative of downstream tasks to evaluate the effectiveness of inpainting. For scene text images, we engage three recognizers, namely CRNN~\cite{shi2016end}, ASTER~\cite{shi2018aster}, and MORAN~\cite{luo2019moran}. These recognizers are well-regarded in the field of scene text image processing~\cite{wang2020scene} and are used to evaluate word-level recognition accuracy (\%). 
On the other hand, when dealing with handwritten text images, we turn to two user-friendly, open-source methods: DAN~\cite{wang2020decoupled} and two versions of~\cite{li2023trocr}—TrOCR-Base and TrOCR-Large. These methods release official weightings and gauge the same metric as applied to scene text images.

In conclusion, our proposed datasets enjoy three characteristics: (1) They cater to the challenges of inpainting both scene text and handwritten texts. (2) Rather than solely relying on synthetic images, we collect images from real-life scenarios for testing, accompanied by the design of realistic and varied forms of corrosion. (3) Beyond the general inpainting task, we evaluate the text image inpainting task via improvement on downstream tasks and image quality.

\section{Methodology}

This section initially provides an overview of the proposed Global Structure-guided Diffusion Model (GSDM). Subsequently, we delve into a detailed explanation of the two units within GSDM: the Structure Prediction Module (SPM) and the Reconstruction Module (RM).

\subsection{Overall Architecture}

The overall architecture of the proposed GSDM is depicted in Figure~\ref{fig:archi}. For the input corrupted text image $\boldsymbol{c}\in \mathbb{R}^{h \times w \times c}$, the SPM first predicts the complete global structure $\boldsymbol{\tilde{s}}\in \mathbb{R}^{h \times w}$. Subsequently, the diffusion-based RM, taking $\boldsymbol{c}$ and $\boldsymbol{\tilde{s}}$ as conditions, generate the intact text image $\boldsymbol{\tilde{x}}\in \mathbb{R}^{h \times w \times c}$.  

\subsection{Structure Prediction Module}
In practice, the content uncertainties in text images are dominated by the global structures, specifically the segmentation mask, of the foreground~\cite{zhu2023dpmn}. Consequently, our aim is to obtain a global structure that closely resembles the original intact image, thereby guiding the subsequent diffusion models in reconstructing corrupted images. To address this challenge, we propose the Structure Prediction Module (SPM), which utilizes a single U-Net~\cite{ronneberger2015u} to predict the correct foreground segmentation masks of intact images via the corrupted ones.

As depicted in Figure~\ref{fig:archi}(b), we utilize a compact U-Net~\cite{ronneberger2015u} denoted as $g_{\theta}$, with three pairs of symmetrical residual blocks to predict the complete segmentation map. Notably, to increase the receptive field and enhance the perception of surrounding corrupted regions, we incorporate dilated convolution~\cite{yu2017dilated} into the network. The prediction process can be formulated as: $\tilde{\boldsymbol{s}}=g_{\theta}(\boldsymbol{c})$.

Given the inherent difficulty of one-stage segmentation prediction, we employ multiple loss functions to compare the actual segmentation map $\boldsymbol{s}$ and the predicted one $\tilde{\boldsymbol{s}}$. Specifically, we implement pixel-level Mean Absolute Error (MAE) loss $\mathcal{L} _{pix}$ and binary segmentation loss $\mathcal{L} _{seg}$ to ensure accurate 2-D segmentation mask generation. The equations are as follows:
\begin{equation}
\mathcal{L} _{pix}=||\boldsymbol{s}-\tilde{\boldsymbol{s}}||_{1},
\label{eq:l_pix}
\end{equation}
\begin{equation}
\mathcal{L} _{seg}=-\frac{1}{N}\sum_{i=1}^N({2 \cdot \boldsymbol{s}_i}\log \tilde{\boldsymbol{s}}_i +{(1-\boldsymbol{s}_i)}\log (1-\tilde{\boldsymbol{s}}_i)),
\label{eq:l_seg}
\end{equation}
where $N$ represents the total number of pixels in an image.

Alternatively, we formulate the character perceptual loss $\mathcal{L} _{cha}$ and style loss $\mathcal{L} _{sty}$ to maintain semantic consistency. We utilize the preamble perceptual layers $\phi _{Rec}$ of a pre-trained text recognizer~\cite{shi2016end} to obtain the feature maps, which are then constrained by the MAE loss. This operation, unlike previous work~\cite{wang2018high}, can effectively capture the semantics of text within the image. The two loss functions are defined as follows:
\begin{equation}
\mathcal{L} _{cha}=||\phi _{Rec}\left( \boldsymbol{s} \right) -\phi _{Rec}\left( \tilde{\boldsymbol{s}} \right) ||_1,
\label{eq:l_tp}
\end{equation}
\begin{equation}
\mathcal{L} _{sty}=||\mathrm{Gram}\left( \boldsymbol{s} \right) -\mathrm{Gram}\left( \tilde{\boldsymbol{s}} \right) ||_1,
\label{eq:l_st}
\end{equation}
where $\mathrm{Gram}$ represents the Gram matrix~\cite{gatys2015neural}. Therefore, the total optimization objective of SPM can be formulated as:
\begin{equation}
\mathcal{L} _{spm}=\lambda _1\mathcal{L} _{pix} + \lambda _2\mathcal{L} _{seg}+\lambda _3\mathcal{L} _{cha}+\lambda _4\mathcal{L} _{sty}.
\label{eq:l_s1}
\end{equation}

\subsection{Reconstruction Module}

Previous diffusion-based inpainting methods~\cite{lugmayr2022repaint,ji2023improving,fei2023generative} rely on the known mask of corrupted regions. In contrast, our model leverages the predicted global structure and corrupted image as conditions to generate an intact text image. Meanwhile, our diffusion model is implemented by vanilla U-Net~\cite{ronneberger2015u} with five pairs of symmetrical residual blocks (Shown in Figure~\ref{fig:archi} (c)). 

\subsubsection{Training Procedure}

As evidenced in~\cite{song2020denoising}, the optimization objective of DDIM is equivalent to the vanilla DDPM. Hence, we adopt the training procedure of the latter.
Given the intact text image $\boldsymbol{x}^{gt}$ as $\boldsymbol{x}_{0}$, we successively add Gaussian noise $\boldsymbol{\epsilon}$ based on the time step $t$, as follows:
\begin{equation}
q(\boldsymbol{x}_{t}|\boldsymbol{x}_{t-1})=\mathcal{N}(\boldsymbol{x}_{t};\sqrt{\alpha_{t}}\boldsymbol{x}_{t-1},(1-\alpha_{t})\boldsymbol{I}),
\label{eq:1}
\end{equation}
where $\alpha_{t}$ is a hyper-parameter between 0 and 1. With the assistance of the reparameterization trick~\cite{ho2020denoising}, the process can be expressed in a more general form:
\begin{equation}
\boldsymbol{x}_{t} = \sqrt{\bar{\alpha}_{t}}\boldsymbol{x}_{0} + \sqrt{1-\bar{\alpha}_{t}}\boldsymbol{\epsilon}, 
\label{eq:2}
\end{equation}
where $\boldsymbol{\epsilon} \sim \mathcal{N}(\boldsymbol{0},\boldsymbol{I})$ and $\bar{\alpha}_{t}=\prod^{t}_{i=0}{\alpha_{i}}\in[0,1]$. 

Following the noise-adding process, we adopt the methodology of DALL-E 2~\cite{ramesh2022hierarchical,xia2023diffir}, which predicts the target image rather than the noise, to improve performance (See ablation study for details). Concretely, receiving the corrupted text image $\boldsymbol{c}$ and predicted segmentation mask $\boldsymbol{\tilde{s}}$ as conditions, the denoising process can be formulated as:
\begin{equation}
p_{f_{\theta}}(\boldsymbol{x}_{t-1}|\boldsymbol{x}_{t},\boldsymbol{c},\tilde{\boldsymbol{s}})=q(\boldsymbol{x}_{t-1}|\boldsymbol{x}_{t},f_{\theta}\left( \boldsymbol{x}_t, \boldsymbol{c},\tilde{\boldsymbol{s}} ,t \right)),
\label{eq:3}
\end{equation}
where $\boldsymbol{c}$ and $\tilde{\boldsymbol{s}}$ are the conditions. Notably, these conditions are concatenated with $\boldsymbol{x}_{t}$ in each step. The total process is supervised by the MSE loss, as:
\begin{equation}
\mathcal{L} _{rm}=||\boldsymbol{x}_o-f_{\theta}\left( \boldsymbol{x}_t, \boldsymbol{c},\tilde{\boldsymbol{s}} ,t \right) ||_2.
\label{eq:ls2}
\end{equation}

\subsubsection{Inference Procedure}
The vanilla DDPM~\cite{ho2020denoising} is time-consuming due to the large number of sampling steps required to maintain high-quality generation. 
During the inference procedure, we perform a non-Markov process~\cite{song2020denoising} to accelerate inference and enhance efficiency. Assuming the original generation sequence is $L=[T,T-1,...,1]$, where the total number of generation steps is $T$, we can construct a sub-sequence $\tau=[\tau_{s},\tau_{s-1},...,1]$ for inference, and the step number is $S \ll T$. The final reconstruction result $\boldsymbol{\tilde{x}}$, can be achieved after $S$ steps, where each step can be written as:
\begin{equation}
\begin{aligned}
\boldsymbol{x}_{\tau_{s-1}}=f_{\theta}\left( \boldsymbol{x}_{\tau_{s}}, \boldsymbol{c},\tilde{\boldsymbol{s}} ,{\tau_{s}} \right).
\end{aligned}
\label{eq:4}
\end{equation}

\section{Experiments}

In this section, we conduct comparison experiments and ablation studies to demonstrate the superiority of our method. Meanwhile, one potential downstream application is presented to show the significance of our work. 

\subsection{Comparison with State-of-the-Art Approaches}

\begin{figure*}[ht]
\centering    
\includegraphics[width=1.0\linewidth]{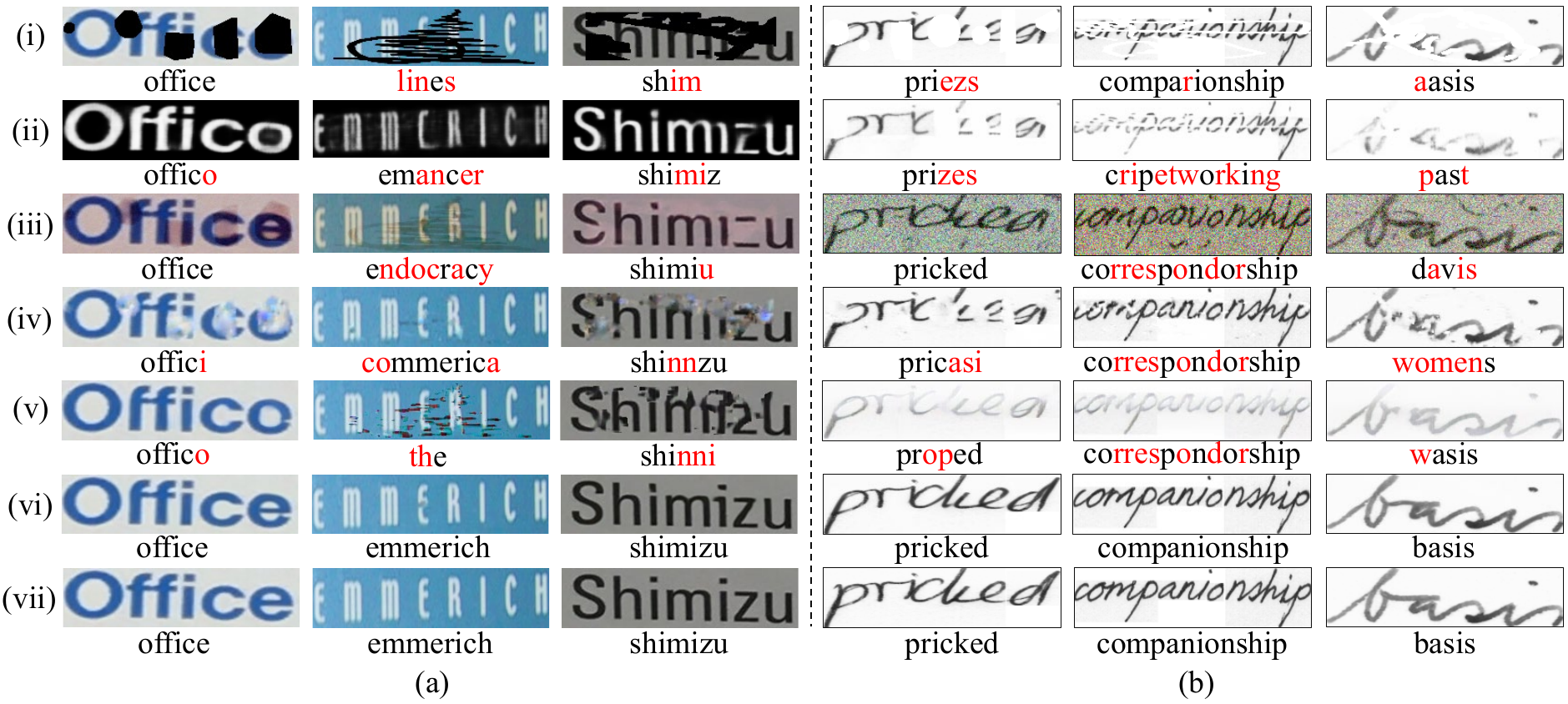}
\caption{The inpainting images with recognition results on TII-ST (ASTER) and TII-HT (TrOCR-L). Red characters indicate errors. The (i) to (vii) denote Corrupted Images, TSINIT/Wang et al., DDIM, CoPaint, TransCNN, GSDM, and GT, respectively.}
\label{fig:comparison}
\end{figure*}

\subsubsection{Scene Text Image}
In this section, we benchmark our proposed approach against prominent existing methods. We first examine the vanilla conditional DDIM~\cite{song2020denoising} and two notable inpainting techniques: TransCNN-HAE~\cite{zhao2022transcnn} (abbr. TransCNN) and CoPaint~\cite{zhang2023towards}. Notably, as a non-blind diffusion-based model, CoPaint can obtain the corrupted masks of each testing image. Additionally, we draw comparisons with the relational technique TSINIT~\cite{sun2022tsinit}, which is designed for binary foreground text completion. As evident from Table~\ref{tab:comparision_ST}, our proposed GSDM outperforms other methods in terms of both recognition accuracy and image quality. 
Notably, our method surpasses both blind and non-blind state-of-the-art methods, i.e., TransCNN and CoPaint.
\begin{table}[t]
\resizebox{1\linewidth}{!}{
{\LARGE
\begin{tabular}{l|ccccc}
\Xhline{1.2pt}
Dataset & \multicolumn{5}{c}{TII-ST} \\ \hline
Metric & CRNN & ASTER & \multicolumn{1}{c|}{MORAN} & PSNR & SSIM \\ \hline
Corrupted Image & 16.89 & 26.21 & \multicolumn{1}{c|}{27.08} & 14.24 & 0.7018 \\ \hline
TSINIT \dag & 56.54 & 63.60 & \multicolumn{1}{c|}{61.22} & - & - \\ 
DDIM & 50.59 & 60.73 &  \multicolumn{1}{c|}{58.53} & 16.79 & 0.7007 \\
CoPaint* & 56.91 & 66.23 & \multicolumn{1}{c|}{65.73} & 26.21 & 0.8794 \\
TransCNN-HAE & 60.41 & 70.61 & \multicolumn{1}{c|}{70.55} & 28.36 & 0.9164 \\
GSDM (ours) & \textbf{67.48} & \textbf{74.67} & \multicolumn{1}{c|}{\textbf{73.04}} & \textbf{33.28} & \textbf{0.9596} \\ \hline
Ground Truth & 80.18 & 88.74 & \multicolumn{1}{c|}{86.93} & - & - \\ \Xhline{1.2pt}
\end{tabular}}}
\caption{The comparison results on TII-ST. The ``-" denote unavailable. ``*" and ``\dag" denote the non-blind method and reproduction version by ourselves, respectively.}
  \label{tab:comparision_ST}
\end{table}
Furthermore, visualization examples from TII-ST can be seen in Figure~\ref{fig:comparison}(a). Two key observations can be made: (1) While some comparison methods may produce correct recognition results, the recovered images often lack style consistency. In contrast, our GSDM ensures not only correct recognition results but also a harmonious and visually appealing style. (2) Ambiguous corrupted regions in images, such as the ``e" in the word ``office", tend to misguide comparison methods into generating incorrect characters. Conversely, our GSDM consistently generates words that are syntactically accurate.

\begin{table}[t]
\resizebox{1\linewidth}{!}{
{\huge
\begin{tabular}{l|ccccc}
\Xhline{1.2pt}
Dataset & \multicolumn{5}{c}{TII-HT} \\ \hline
Metric & DAN & TrOCR-B & \multicolumn{1}{c|}{TrOCR-L} & PSNR & SSIM \\ \hline
Corrupted Image & 23.81 & 19.75 & \multicolumn{1}{c|}{33.25} & 20.08 & 0.8916 \\ \hline
Wang et al. \dag & 21.63 & 11.00 & \multicolumn{1}{c|}{18.50} & 16.89 & 0.8113 \\ 
DDIM & 0.25 & 10.75 & \multicolumn{1}{c|}{44.13} & 9.32 & 0.2842 \\
CoPaint* & 42.12 & 26.06 & \multicolumn{1}{c|}{45.50} & 24.52 & 0.9203 \\
TransCNN-HAE & 17.19 & 22.87 & \multicolumn{1}{c|}{47.25} & 15.42 & 0.7675 \\
GSDM (ours) & \textbf{69.43} & \textbf{56.00} & \multicolumn{1}{c|}{\textbf{66.81}} & \textbf{32.13} & \textbf{0.9718} \\ \hline
Ground Truth & 85.19 & 64.07 & \multicolumn{1}{c|}{75.56} & - & - \\ \Xhline{1.2pt}
\end{tabular}}}
\caption{The comparison results on TII-HT. The ``-" denote unavailable. ``*" and ``\dag" denote the non-blind method and reproduction version by ourselves, respectively.}
  \label{tab:comparison_HT}
\end{table}

\subsubsection{Handwritten Text Image}

In evaluating handwritten text images, we maintain the aforementioned comparison methods but substitute TSINIT with a character inpainting one~\cite{wang2021chinese} (Reproduced and modified for this task). 
As depicted in Table~\ref{tab:comparison_HT}, our methods achieve pleasing performance in terms of both recognition accuracy and image quality. Figure~\ref{fig:comparison}(b) reveals that our approach is adept at delicately restoring the strokes. In stark contrast, comparison methods manifest varying levels of quality degradation, leading to unstable recognition accuracy. Notably, although CoPaint can generate visually appealing images, its recognition outcomes are often erroneous. This can be attributed to the fact that HTR methods are sensitive to structural completeness. That is, even minor corrosion can mislead recognizers, resulting in incorrect outputs.

\subsection{Ablation Study}
Here we delve into the impact of various components within our proposed method. To maintain consistency, all experiments are conducted on the scene text image dataset, TII-ST. The recognition accuracy represents the average results derived from CRNN, ASTER, and MORAN.

\subsubsection{Variants of the GSDM}
In this study, we investigate the significance of different components within our GSDM. To do this, we directly applied different components to reconstruct the corrupted text images. The results, presented in Table~\ref{tab:GlobalStructure}, reveal the following insights: (1) The standalone SPM yields trivial results, attributable to the inherent limitations of the traditional U-Net model in generating diverse text image styles. (2) GSDM surpasses a singular reconstruction module, underscoring the benefits of integrating a global structure. (3) Compared to traditional noise-predicting diffusion methods, predicting the image denoted by $\boldsymbol{x}$ emerges as significantly superior. 
A plausible reason behind this is the robustness introduced by this paradigm during training.

\begin{table}[t]
\centering
\begin{tabular}{cc|ccc}
\Xhline{1.2pt}
Architecture & Target & Accuracy & PSNR & SSIM \\ \hline
SPM & - & 66.59 & 25.90 & 0.8722 \\ \hline
\multirow{2}{*}{RM} & $\boldsymbol{\epsilon}$ & 55.35 & 16.79 &  0.7007 \\
 & $\boldsymbol{x}$ & 69.40 & 32.59 & 0.9561 \\ \hline
\multirow{2}{*}{SPM+RM} & $\boldsymbol{\epsilon}$ & 56.10 & 16.72 & 0.7112 \\ 
 & $\boldsymbol{x}$ (ours) & \textbf{71.73} & \textbf{33.28} & \textbf{0.9596}  \\ \Xhline{1.2pt}
\end{tabular}
\caption{The performance of different architecture.}
\label{tab:GlobalStructure}
\end{table}

\subsubsection{Effect of Sampling Strategy in RM}
We conduct experiments to demonstrate the efficacy of the chosen sampling strategy in the RM. Results in Table~\ref{tab:sampling} show that: (1) By adopting the Non-Markov strategy, inspired by DDIM~\cite{song2020denoising}, our proposed method significantly outperforms its Markov-strategy counterpart from the vanilla DDPM~\cite{ho2020denoising}, in terms of both performance metrics and computational efficiency. (2) We observe a noticeable drop in performance as the number of inference steps increases in our approach. One possible explanation is that while our method generates high-quality images in a single step, repeated regeneration of the target image introduces noise cumulatively. 

\begin{table}[ht]
\centering
\resizebox{1\linewidth}{!}{
{\huge
\begin{tabular}{c|c|cccc}
\Xhline{1.2pt}
Strategy & Step & Accuracy & PSNR & SSIM & Time (s) \\ \hline
\multirow{3}{*}{Markov} & 100 & 66.23 & 30.51 & 0.9386 & 1.720 \\
 & 500 & 68.35 & 32.05 & 0.9535 & 8.670 \\
 & 1000 & 68.21 & 32.28 & 0.9401 & 17.560 \\ \hline
\multirow{3}{*}{Non-Markov} & 1 & \textbf{71.73} & \textbf{33.28} & \textbf{0.9596} & \textbf{0.034} \\
 & 5 & 69.38 & 33.03 & 0.9582 & 0.110 \\
 & 10 & 68.96 & 32.87 & 0.9575 & 0.250 \\ \Xhline{1.2pt}
\end{tabular}}}
\caption{Performance of   in our reconstruction module. The ``Step" indicates the number of sampling steps during inference.}
\label{tab:sampling}
\end{table}

\subsubsection{Effect of the Training Objective}
In this study, we investigate the training objective of the proposed GSDM. It is noted that the baseline is primarily optimized by $\mathcal{L}_{pix}$ and $\mathcal{L}_{rm}$. The results in Table~\ref{tab:loss} show that: (1) Even when constrained by basic loss functions, our baseline demonstrates superior recognition performance compared to the state-of-the-art blind method~\cite{zhao2022transcnn} (69.80 vs. 67.19). (2) The recognition performance of GSDM is significantly improved by including more types of loss functions. Notably, the synergistic optimization effect of $\mathcal{L}_{cha}$ and $\mathcal{L}_{sty}$, which aim to maintain semantic consistency, greatly outperforms each of them (See (iii)--(v)). (3) Unlike the improvement in recognition performance, there is no significant change in image quality. This may be attributed to two factors. On one hand, our robust diffusion-based baseline is capable of producing high-quality images. On the other hand, all these loss functions are exerted on SPM, enabling RM to generate more accurate text content.

\begin{table}[ht]
\centering
\resizebox{1\linewidth}{!}{
{\huge
\begin{tabular}{c|cccc|ccc}
\Xhline{1.2pt}
& Baseline &  $\mathcal{L}_{seg}$ & $\mathcal{L}_{cha}$ & $\mathcal{L}_{sty}$ & Accuracy & PSNR & SSIM \\ \hline
(i) & \checkmark & & & & 69.80 & 33.34 & 0.9600 \\
(ii) & \checkmark & \checkmark & & &  70.00 & 33.32 & 0.9598 \\
(iii) & \checkmark & \checkmark & \checkmark & & 70.19 & 33.30 & 0.9598 \\
(iv) & \checkmark & \checkmark &  & \checkmark & 70.09 & 33.31 & 0.9598 \\

(v) & \checkmark & \checkmark & \checkmark & \checkmark & 71.73 & 33.28 & 0.9596 \\ \Xhline{1.2pt}
\end{tabular}}}
\caption{The performance of different training objectives.}
\label{tab:loss}
\end{table}

\subsection{Improvement on Scene Text Editing}

\begin{figure}[t]
\centering
\includegraphics[width=1\columnwidth]{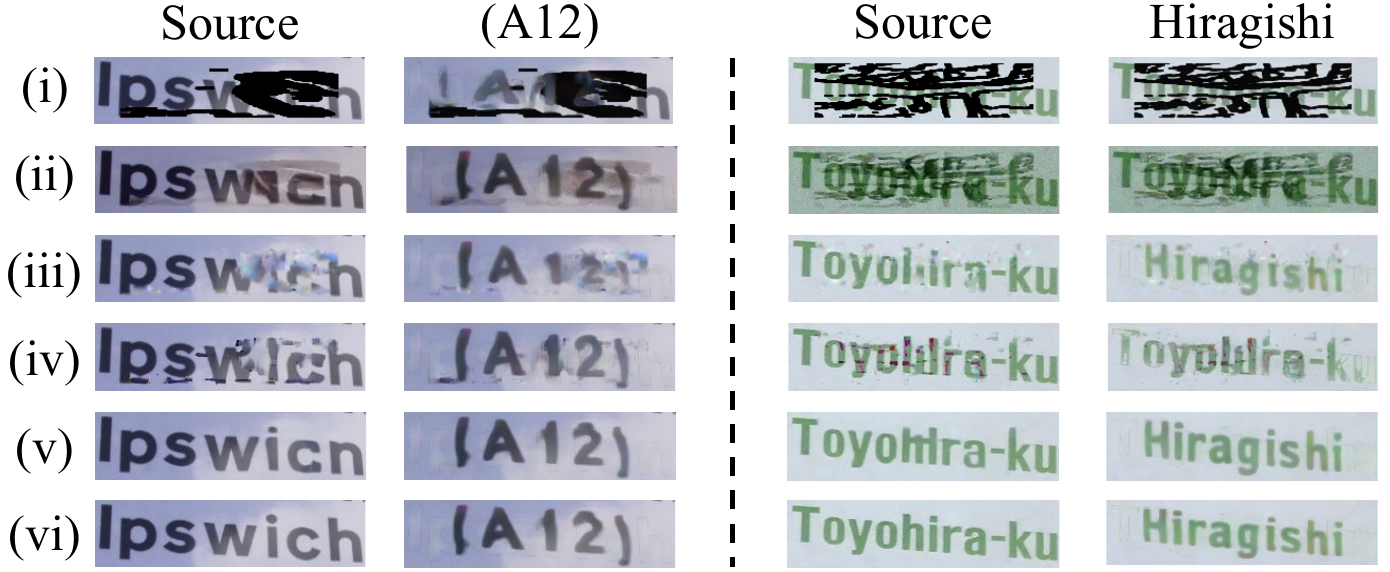} 
\caption{Influence of inpainting methods on scene text image editing. The ``Source'' denotes the source image. ``(A12)'' and ``Hiragishi'' denotes the guidance texts. The (i) to (vi) denote Corrupted images, DDIM, CoPaint, TransCNN, GSDM, and GT, respectively.}
\label{fig:application_STE}
\end{figure}

To further evaluate the improvement of text inpainting tasks in downstream applications, we conduct a preliminary experiment on scene text editing. This task involves replacing text within a scene image with new content while preserving the original style, as described in~\cite{wu2019editing}. Such an approach has proven invaluable in real-world applications, including augmented reality translation. We adopted the recent MOSTEL framework~\cite{qu2023exploring} to demonstrate the significance of our task. As shown in Figure~\ref{fig:application_STE}, edits made on corrupted images are often unsatisfactory. In addition, the subpar inpainting performance of several comparison methods introduces artifacts into the text editing process. Some methods, such as DDIM, generate images that MOSTEL struggles to model effectively. In contrast, the repaired images from our proposed GSDM model yield consistently high-quality results, comparable to those from unaltered images. This finding underscores the importance of prioritizing image quality in inpainting tasks.

\section{Conclusion}

Given the observation of corrosion issues in real-world text, we study a new task: text image inpainting, aiming to repair corrupted images. 
To this end, we develop two datasets tailored for the target task, namely TII-ST and TII-HT. Concurrently, a novel approach, the Global Structure-guided Diffusion Model (GSDM), is proposed to fulfill text inpainting. Although text image inpainting is a challenging task, comprehensive experiments verify the effectiveness of our method, which enhances both image quality and the performance of the downstream recognition task. We believe the proposed task in this paper introduces a new branch for image inpainting, which will pose considerable significance in repairing text images in real-world scenarios. Future studies include improving the inpainting performance and exploring the applications that benefited from the proposed task.

\section{Acknowledgments}
This work was supported by the National Natural Science Foundation of China (Nos. 62076062 and 62306070) and the Social Development Science and Technology Project of Jiangsu Province (No. BE2022811). Furthermore, the work was also supported by the Big Data Computing Center of Southeast University. Thanks for the help of three interns, Bihong Wang, Chenxing Liu, and Tianxu Li. 
\bibliography{aaai24}

\begin{thebibliography}{59}
\providecommand{\natexlab}[1]{#1}

\bibitem[{Bautista and Atienza(2022)}]{bautista2022parseq}
Bautista, D.; and Atienza, R. 2022.
\newblock Scene Text Recognition with Permuted Autoregressive Sequence Models.
\newblock In \emph{Proceedings of the European Conference on Computer Vision}, 178--196.

\bibitem[{Bertalmio et~al.(2003)Bertalmio, Vese, Sapiro, and Osher}]{bertalmio2003simultaneous}
Bertalmio, M.; Vese, L.; Sapiro, G.; and Osher, S. 2003.
\newblock Simultaneous structure and texture image inpainting.
\newblock \emph{IEEE Transactions on Image Processing}, 12(8): 882--889.

\bibitem[{Cai et~al.(2017)Cai, Su, Lin, Wang, Yang, and Ling}]{cai2017blind}
Cai, N.; Su, Z.; Lin, Z.; Wang, H.; Yang, Z.; and Ling, B. W.-K. 2017.
\newblock Blind inpainting using the fully convolutional neural network.
\newblock \emph{The Visual Computer}, 33: 249--261.

\bibitem[{Chang et~al.(2018)Chang, Gu, Zhang, Wang, and Innovation}]{chang2018chinese}
Chang, J.; Gu, Y.; Zhang, Y.; Wang, Y.-F.; and Innovation, C. 2018.
\newblock Chinese Handwriting Imitation with Hierarchical Generative Adversarial Network.
\newblock In \emph{Proceedings of the British Machine Vision Conference}, 290.

\bibitem[{Devlin et~al.(2018)Devlin, Chang, Lee, and Toutanova}]{devlin2018bert}
Devlin, J.; Chang, M.-W.; Lee, K.; and Toutanova, K. 2018.
\newblock Bert: Pre-training of deep bidirectional transformers for language understanding.
\newblock \emph{arXiv preprint arXiv:1810.04805}.

\bibitem[{Fang et~al.(2021)Fang, Xie, Wang, Mao, and Zhang}]{fang2021read}
Fang, S.; Xie, H.; Wang, Y.; Mao, Z.; and Zhang, Y. 2021.
\newblock Read like humans: Autonomous, bidirectional and iterative language modeling for scene text recognition.
\newblock In \emph{Proceedings of the IEEE/CVF Conference on Computer Vision and Pattern Recognition}, 7098--7107.

\bibitem[{Fei et~al.(2023)Fei, Lyu, Pan, Zhang, Yang, Luo, Zhang, and Dai}]{fei2023generative}
Fei, B.; Lyu, Z.; Pan, L.; Zhang, J.; Yang, W.; Luo, T.; Zhang, B.; and Dai, B. 2023.
\newblock Generative Diffusion Prior for Unified Image Restoration and Enhancement.
\newblock In \emph{Proceedings of the IEEE/CVF Conference on Computer Vision and Pattern Recognition}, 9935--9946.

\bibitem[{Gatys, Ecker, and Bethge(2015)}]{gatys2015neural}
Gatys, L.~A.; Ecker, A.~S.; and Bethge, M. 2015.
\newblock A neural algorithm of artistic style.
\newblock \emph{arXiv preprint arXiv:1508.06576}.

\bibitem[{Graham(1972)}]{graham1972efficient}
Graham, R.~L. 1972.
\newblock An efficient algorithm for determining the convex hull of a finite planar set.
\newblock \emph{Information Processing Letter}, 1: 132--133.

\bibitem[{Gupta, Vedaldi, and Zisserman(2016)}]{gupta2016synthetic}
Gupta, A.; Vedaldi, A.; and Zisserman, A. 2016.
\newblock Synthetic data for text localisation in natural images.
\newblock In \emph{Proceedings of the IEEE Conference on Computer Vision and Pattern Recognition}, 2315--2324.

\bibitem[{He et~al.(2023)He, Wang, Hu, Liu, Liu, Xu, and Shen}]{he2023icl}
He, J.; Wang, L.; Hu, Y.; Liu, N.; Liu, H.; Xu, X.; and Shen, H.~T. 2023.
\newblock ICL-D3IE: In-context learning with diverse demonstrations updating for document information extraction.
\newblock \emph{arXiv preprint arXiv:2303.05063}.

\bibitem[{Ho, Jain, and Abbeel(2020)}]{ho2020denoising}
Ho, J.; Jain, A.; and Abbeel, P. 2020.
\newblock Denoising diffusion probabilistic models.
\newblock \emph{Proceedings of the Advances in Neural Information Processing Systems}, 33: 6840--6851.

\bibitem[{Jaderberg et~al.(2014)Jaderberg, Simonyan, Vedaldi, and Zisserman}]{jaderberg2014synthetic}
Jaderberg, M.; Simonyan, K.; Vedaldi, A.; and Zisserman, A. 2014.
\newblock Synthetic data and artificial neural networks for natural scene text recognition.
\newblock \emph{arXiv preprint arXiv:1406.2227}.

\bibitem[{Ji et~al.(2023)Ji, Zhang, Wang, Hou, Zhang, Price, and Chang}]{ji2023improving}
Ji, J.; Zhang, G.; Wang, Z.; Hou, B.; Zhang, Z.; Price, B.; and Chang, S. 2023.
\newblock Improving Diffusion Models for Scene Text Editing with Dual Encoders.
\newblock \emph{arXiv preprint arXiv:2304.05568}.

\bibitem[{Karatzas et~al.(2015)Karatzas, Gomez-Bigorda, Nicolaou, Ghosh, Bagdanov, Iwamura, Matas, Neumann, Chandrasekhar, Lu et~al.}]{karatzas2015icdar}
Karatzas, D.; Gomez-Bigorda, L.; Nicolaou, A.; Ghosh, S.; Bagdanov, A.; Iwamura, M.; Matas, J.; Neumann, L.; Chandrasekhar, V.~R.; Lu, S.; et~al. 2015.
\newblock ICDAR 2015 competition on robust reading.
\newblock In \emph{Proceedings of the International Conference on Document Analysis and Recognition}, 1156--1160.

\bibitem[{Karatzas et~al.(2013)Karatzas, Shafait, Uchida, Iwamura, i~Bigorda, Mestre, Mas, Mota, Almazan, and De~Las~Heras}]{karatzas2013icdar}
Karatzas, D.; Shafait, F.; Uchida, S.; Iwamura, M.; i~Bigorda, L.~G.; Mestre, S.~R.; Mas, J.; Mota, D.~F.; Almazan, J.~A.; and De~Las~Heras, L.~P. 2013.
\newblock ICDAR 2013 robust reading competition.
\newblock In \emph{Proceedings of the International Conference on Document Analysis and Recognition}, 1484--1493.

\bibitem[{Kawar et~al.(2022)Kawar, Elad, Ermon, and Song}]{kawar2022denoising}
Kawar, B.; Elad, M.; Ermon, S.; and Song, J. 2022.
\newblock Denoising diffusion restoration models.
\newblock \emph{Advances in Neural Information Processing Systems}, 35: 23593--23606.

\bibitem[{Kingma and Ba(2014)}]{kingma2014adam}
Kingma, D.~P.; and Ba, J. 2014.
\newblock Adam: A method for stochastic optimization.
\newblock \emph{arXiv preprint arXiv:1412.6980}.

\bibitem[{Kirillov et~al.(2023)Kirillov, Mintun, Ravi, Mao, Rolland, Gustafson, Xiao, Whitehead, Berg, Lo et~al.}]{kirillov2023segment}
Kirillov, A.; Mintun, E.; Ravi, N.; Mao, H.; Rolland, C.; Gustafson, L.; Xiao, T.; Whitehead, S.; Berg, A.~C.; Lo, W.-Y.; et~al. 2023.
\newblock Segment anything.
\newblock \emph{arXiv preprint arXiv:2304.02643}.

\bibitem[{Krishnan et~al.(2023)Krishnan, Kovvuri, Pang, Vassilev, and Hassner}]{krishnan2023textstylebrush}
Krishnan, P.; Kovvuri, R.; Pang, G.; Vassilev, B.; and Hassner, T. 2023.
\newblock TextStyleBrush: Transfer of Text Aesthetics From a Single Example.
\newblock \emph{IEEE Transactions on Pattern Analysis and Machine Intelligence}, 45(7): 9122--9134.

\bibitem[{Lai et~al.(2021)Lai, Jin, Zhu, Li, and Lin}]{lai2021synsig2vec}
Lai, S.; Jin, L.; Zhu, Y.; Li, Z.; and Lin, L. 2021.
\newblock SynSig2Vec: Forgery-free learning of dynamic signature representations by sigma lognormal-based synthesis and 1D CNN.
\newblock \emph{IEEE Transactions on Pattern Analysis and Machine Intelligence}, 44(10): 6472--6485.

\bibitem[{Levenshtein et~al.(1966)}]{levenshtein1966binary}
Levenshtein, V.~I.; et~al. 1966.
\newblock Binary codes capable of correcting deletions, insertions, and reversals.
\newblock In \emph{Soviet physics doklady}, volume~10, 707--710.

\bibitem[{Li et~al.(2023)Li, Lv, Chen, Cui, Lu, Florencio, Zhang, Li, and Wei}]{li2023trocr}
Li, M.; Lv, T.; Chen, J.; Cui, L.; Lu, Y.; Florencio, D.; Zhang, C.; Li, Z.; and Wei, F. 2023.
\newblock Trocr: Transformer-based optical character recognition with pre-trained models.
\newblock In \emph{Proceedings of the AAAI Conference on Artificial Intelligence}, 13094--13102.

\bibitem[{Liu et~al.(2018)Liu, Reda, Shih, Wang, Tao, and Catanzaro}]{liu2018image}
Liu, G.; Reda, F.~A.; Shih, K.~J.; Wang, T.-C.; Tao, A.; and Catanzaro, B. 2018.
\newblock Image inpainting for irregular holes using partial convolutions.
\newblock In \emph{Proceedings of the European Conference on Computer Vision}, 85--100.

\bibitem[{Long, He, and Yao(2021)}]{long2021scene}
Long, S.; He, X.; and Yao, C. 2021.
\newblock Scene text detection and recognition: The deep learning era.
\newblock \emph{International Journal of Computer Vision}, 129: 161--184.

\bibitem[{Lugmayr et~al.(2022)Lugmayr, Danelljan, Romero, Yu, Timofte, and Van~Gool}]{lugmayr2022repaint}
Lugmayr, A.; Danelljan, M.; Romero, A.; Yu, F.; Timofte, R.; and Van~Gool, L. 2022.
\newblock Repaint: Inpainting using denoising diffusion probabilistic models.
\newblock In \emph{Proceedings of the IEEE/CVF Conference on Computer Vision and Pattern Recognition}, 11461--11471.

\bibitem[{Luo, Jin, and Sun(2019)}]{luo2019moran}
Luo, C.; Jin, L.; and Sun, Z. 2019.
\newblock Moran: A multi-object rectified attention network for scene text recognition.
\newblock \emph{Pattern Recognition}, 90: 109--118.

\bibitem[{Marti and Bunke(2002)}]{marti2002iam}
Marti, U.-V.; and Bunke, H. 2002.
\newblock The IAM-database: an English sentence database for offline handwriting recognition.
\newblock \emph{International Journal on Document Analysis and Recognition}, 5: 39--46.

\bibitem[{Nayef et~al.(2017)Nayef, Yin, Bizid, Choi, Feng, Karatzas, Luo, Pal, Rigaud, Chazalon et~al.}]{nayef2017icdar2017}
Nayef, N.; Yin, F.; Bizid, I.; Choi, H.; Feng, Y.; Karatzas, D.; Luo, Z.; Pal, U.; Rigaud, C.; Chazalon, J.; et~al. 2017.
\newblock Icdar2017 robust reading challenge on multi-lingual scene text detection and script identification-rrc-mlt.
\newblock In \emph{Proceedings of the International Conference on Document Analysis and Recognition}, 1454--1459.

\bibitem[{Pan et~al.(2021)Pan, Zhan, Dai, Lin, Loy, and Luo}]{pan2021exploiting}
Pan, X.; Zhan, X.; Dai, B.; Lin, D.; Loy, C.~C.; and Luo, P. 2021.
\newblock Exploiting deep generative prior for versatile image restoration and manipulation.
\newblock \emph{IEEE Transactions on Pattern Analysis and Machine Intelligence}, 44(11): 7474--7489.

\bibitem[{Qu et~al.(2023)Qu, Tan, Xie, Xu, Wang, and Zhang}]{qu2023exploring}
Qu, Y.; Tan, Q.; Xie, H.; Xu, J.; Wang, Y.; and Zhang, Y. 2023.
\newblock Exploring stroke-level modifications for scene text editing.
\newblock In \emph{Proceedings of the AAAI Conference on Artificial Intelligence}, 2119--2127.

\bibitem[{Ramesh et~al.(2022)Ramesh, Dhariwal, Nichol, Chu, and Chen}]{ramesh2022hierarchical}
Ramesh, A.; Dhariwal, P.; Nichol, A.; Chu, C.; and Chen, M. 2022.
\newblock Hierarchical text-conditional image generation with clip latents.
\newblock \emph{arXiv preprint arXiv:2204.06125}.

\bibitem[{Rombach et~al.(2022)Rombach, Blattmann, Lorenz, Esser, and Ommer}]{rombach2022high}
Rombach, R.; Blattmann, A.; Lorenz, D.; Esser, P.; and Ommer, B. 2022.
\newblock High-resolution image synthesis with latent diffusion models.
\newblock In \emph{Proceedings of the IEEE/CVF conference on computer vision and pattern recognition}, 10684--10695.

\bibitem[{Ronneberger, Fischer, and Brox(2015)}]{ronneberger2015u}
Ronneberger, O.; Fischer, P.; and Brox, T. 2015.
\newblock U-net: Convolutional networks for biomedical image segmentation.
\newblock In \emph{Proceedings of the International Conference on Medical Image Computing and Computer-Assisted Intervention}, 234--241.

\bibitem[{Shah, Gautam, and Singh(2022)}]{shah2022overview}
Shah, R.; Gautam, A.; and Singh, S.~K. 2022.
\newblock Overview of image inpainting techniques: A survey.
\newblock In \emph{2022 IEEE Region 10 Symposium (TENSYMP)}, 1--6. IEEE.

\bibitem[{Shi, Bai, and Yao(2016)}]{shi2016end}
Shi, B.; Bai, X.; and Yao, C. 2016.
\newblock An end-to-end trainable neural network for image-based sequence recognition and its application to scene text recognition.
\newblock \emph{IEEE Transactions on Pattern Analysis and Machine Intelligence}, 39(11): 2298--2304.

\bibitem[{Shi et~al.(2018)Shi, Yang, Wang, Lyu, Yao, and Bai}]{shi2018aster}
Shi, B.; Yang, M.; Wang, X.; Lyu, P.; Yao, C.; and Bai, X. 2018.
\newblock {ASTER}: An attentional scene text recognizer with flexible rectification.
\newblock \emph{IEEE Transactions on Pattern Analysis and Machine Intelligence}, 41(9): 2035--2048.

\bibitem[{Singh and Karayev(2021)}]{singh2021full}
Singh, S.~S.; and Karayev, S. 2021.
\newblock Full page handwriting recognition via image to sequence extraction.
\newblock In \emph{Proceedings of the International Conference on Document Analysis and Recognition}, 55--69. Springer.

\bibitem[{Song, Meng, and Ermon(2020)}]{song2020denoising}
Song, J.; Meng, C.; and Ermon, S. 2020.
\newblock Denoising diffusion implicit models.
\newblock \emph{arXiv preprint arXiv:2010.02502}.

\bibitem[{Sun et~al.(2022)Sun, Xue, Li, Zhu, Zhang, and Zhang}]{sun2022tsinit}
Sun, J.; Xue, F.; Li, J.; Zhu, L.; Zhang, H.; and Zhang, J. 2022.
\newblock TSINIT: a two-stage Inpainting network for incomplete text.
\newblock \emph{IEEE Transactions on Multimedia}.

\bibitem[{Wan et~al.(2021)Wan, Zhang, Chen, and Liao}]{wan2021high}
Wan, Z.; Zhang, J.; Chen, D.; and Liao, J. 2021.
\newblock High-fidelity pluralistic image completion with transformers.
\newblock In \emph{Proceedings of the IEEE/CVF International Conference on Computer Vision}, 4692--4701.

\bibitem[{Wang et~al.(2021)Wang, Pan, Sun, and Zhang}]{wang2021chinese}
Wang, J.; Pan, G.; Sun, D.; and Zhang, J. 2021.
\newblock Chinese Character Inpainting with Contextual Semantic Constraints.
\newblock In \emph{Proceedings of the 29th ACM International Conference on Multimedia}, 1829--1837.

\bibitem[{Wang, Ouyang, and Chen(2021)}]{wang2021image}
Wang, T.; Ouyang, H.; and Chen, Q. 2021.
\newblock Image Inpainting with External-internal Learning and Monochromic Bottleneck.
\newblock In \emph{Proceedings of the IEEE/CVF Conference on Computer Vision and Pattern Recognition}, 5120--5129.

\bibitem[{Wang et~al.(2020{\natexlab{a}})Wang, Zhu, Jin, Luo, Chen, Wu, Wang, and Cai}]{wang2020decoupled}
Wang, T.; Zhu, Y.; Jin, L.; Luo, C.; Chen, X.; Wu, Y.; Wang, Q.; and Cai, M. 2020{\natexlab{a}}.
\newblock Decoupled attention network for text recognition.
\newblock In \emph{Proceedings of the AAAI conference on artificial intelligence}, 12216--12224.

\bibitem[{Wang et~al.(2018)Wang, Liu, Zhu, Tao, Kautz, and Catanzaro}]{wang2018high}
Wang, T.-C.; Liu, M.-Y.; Zhu, J.-Y.; Tao, A.; Kautz, J.; and Catanzaro, B. 2018.
\newblock High-resolution image synthesis and semantic manipulation with conditional gans.
\newblock In \emph{Proceedings of the IEEE/CVF Conference on Computer Vision and Pattern Recognition}, 8798--8807.

\bibitem[{Wang et~al.(2020{\natexlab{b}})Wang, Xie, Liu, Wang, Liang, Shen, and Bai}]{wang2020scene}
Wang, W.; Xie, E.; Liu, X.; Wang, W.; Liang, D.; Shen, C.; and Bai, X. 2020{\natexlab{b}}.
\newblock Scene text image super-resolution in the wild.
\newblock In \emph{Proceedings of the European Conference on Computer Vision}, 650--666. Springer.

\bibitem[{Wang et~al.(2020{\natexlab{c}})Wang, Chen, Tao, and Jia}]{wang2020vcnet}
Wang, Y.; Chen, Y.-C.; Tao, X.; and Jia, J. 2020{\natexlab{c}}.
\newblock Vcnet: A robust approach to blind image inpainting.
\newblock In \emph{Proceedings of the European Conference on Computer Vision}, 752--768. Springer.

\bibitem[{Wu et~al.(2019)Wu, Zhang, Liu, Han, Liu, Ding, and Bai}]{wu2019editing}
Wu, L.; Zhang, C.; Liu, J.; Han, J.; Liu, J.; Ding, E.; and Bai, X. 2019.
\newblock Editing text in the wild.
\newblock In \emph{Proceedings of the 27th ACM international conference on multimedia}, 1500--1508.

\bibitem[{Xia et~al.(2023)Xia, Zhang, Wang, Wang, Wu, Tian, Yang, and Van~Gool}]{xia2023diffir}
Xia, B.; Zhang, Y.; Wang, S.; Wang, Y.; Wu, X.; Tian, Y.; Yang, W.; and Van~Gool, L. 2023.
\newblock Diffir: Efficient diffusion model for image restoration.
\newblock \emph{arXiv preprint arXiv:2303.09472}.

\bibitem[{Xiang et~al.(2023)Xiang, Zou, Nawaz, Huang, Zhang, and Yu}]{xiang2023deep}
Xiang, H.; Zou, Q.; Nawaz, M.~A.; Huang, X.; Zhang, F.; and Yu, H. 2023.
\newblock Deep learning for image inpainting: A survey.
\newblock \emph{Pattern Recognition}, 134: 109046.

\bibitem[{Yang et~al.(2019)Yang, Dai, Yang, Carbonell, Salakhutdinov, and Le}]{yang2019xlnet}
Yang, Z.; Dai, Z.; Yang, Y.; Carbonell, J.; Salakhutdinov, R.~R.; and Le, Q.~V. 2019.
\newblock Xlnet: Generalized autoregressive pretraining for language understanding.
\newblock \emph{Proceedings of Advances in Neural Information Processing Systems}, 32.

\bibitem[{Yu, Koltun, and Funkhouser(2017)}]{yu2017dilated}
Yu, F.; Koltun, V.; and Funkhouser, T. 2017.
\newblock Dilated residual networks.
\newblock In \emph{Proceedings of the IEEE/CVF Conference on Computer Vision and Pattern Recognition}, 472--480.

\bibitem[{Yu et~al.(2023)Yu, Feng, Feng, Liu, Jin, Zeng, and Chen}]{yu2023inpaint}
Yu, T.; Feng, R.; Feng, R.; Liu, J.; Jin, X.; Zeng, W.; and Chen, Z. 2023.
\newblock Inpaint anything: Segment anything meets image inpainting.
\newblock \emph{arXiv preprint arXiv:2304.06790}.

\bibitem[{Yu et~al.(2022)Yu, Du, Zhang, and Luo}]{yu2022unbiased}
Yu, Y.; Du, D.; Zhang, L.; and Luo, T. 2022.
\newblock Unbiased multi-modality guidance for image inpainting.
\newblock In \emph{Proceedings of the European Conference on Computer Vision}, 668--684. Springer.

\bibitem[{Zhang et~al.(2023)Zhang, Ji, Zhang, Yu, Jaakkola, and Chang}]{zhang2023towards}
Zhang, G.; Ji, J.; Zhang, Y.; Yu, M.; Jaakkola, T.~S.; and Chang, S. 2023.
\newblock Towards Coherent Image Inpainting Using Denoising Diffusion Implicit Models.
\newblock In \emph{Proceedings of the International Conference on Machine Learning}.

\bibitem[{Zhang et~al.(2017)Zhang, He, Sun, and Tan}]{zhang2017demeshnet}
Zhang, S.; He, R.; Sun, Z.; and Tan, T. 2017.
\newblock Demeshnet: Blind face inpainting for deep meshface verification.
\newblock \emph{IEEE Transactions on Information Forensics and Security}, 13(3): 637--647.

\bibitem[{Zhao et~al.(2022)Zhao, Gu, Zheng, and Zheng}]{zhao2022transcnn}
Zhao, H.; Gu, Z.; Zheng, B.; and Zheng, H. 2022.
\newblock Transcnn-hae: Transformer-cnn hybrid autoencoder for blind image inpainting.
\newblock In \emph{Proceedings of the 30th ACM International Conference on Multimedia}, 6813--6821.

\bibitem[{Zhu et~al.(2023{\natexlab{a}})Zhu, Zhao, Fang, and Xue}]{zhu2023dpmn}
Zhu, S.; Zhao, Z.; Fang, P.; and Xue, H. 2023{\natexlab{a}}.
\newblock Improving Scene Text Image Super-resolution via Dual Prior Modulation Network.
\newblock In \emph{Proceedings of the AAAI Conference on Artificial Intelligence}, 3843--3851.

\bibitem[{Zhu et~al.(2023{\natexlab{b}})Zhu, Li, Wang, He, and Yao}]{zhu2023conditional}
Zhu, Y.; Li, Z.; Wang, T.; He, M.; and Yao, C. 2023{\natexlab{b}}.
\newblock Conditional Text Image Generation with Diffusion Models.
\newblock In \emph{Proceedings of the IEEE/CVF Conference on Computer Vision and Pattern Recognition}, 14235--14245.

\end{thebibliography}

\section{APPENDIX}

\section{Details of Benchmark Dataset}

\subsection{Construction Method}

\subsubsection{Data-cleaning Strategy}
\begin{figure}[ht]
\centering
\includegraphics[width=1\columnwidth]{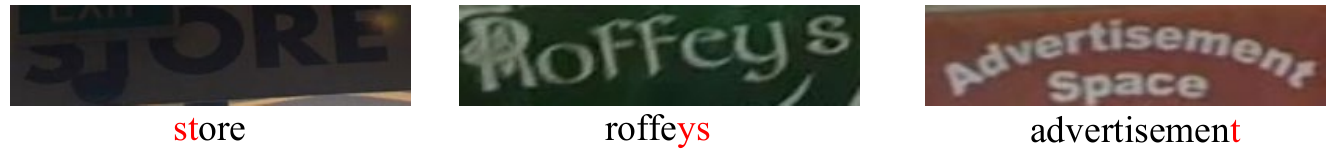} 
\caption{The illustration of excluded samples. The red characters mean the missing ones in the ground truth text labels.}
\label{fig:wrongexamples}
\end{figure}

The objective of text image inpainting is to restore corrupted text images, resulting in enhanced image quality and improved performance in downstream tasks. In real-life datasets such as ICDAR2013~\cite{karatzas2013icdar}, ICDAR2015~\cite{karatzas2015icdar}, and ICDAR2017~\cite{nayef2017icdar2017}, many images exhibit inherent defects, making them unsuitable for our text inpainting tasks. As illustrated in Figure~\ref{fig:wrongexamples}, some samples have mismatching labels, leading to inaccuracies in evaluating downstream recognition tasks. Given this observation, we exclude these problematic samples from the original dataset to create our TII-ST.

\subsubsection{Corrosion Forms}
\begin{figure}[ht]
\centering
\includegraphics[width=1\columnwidth]{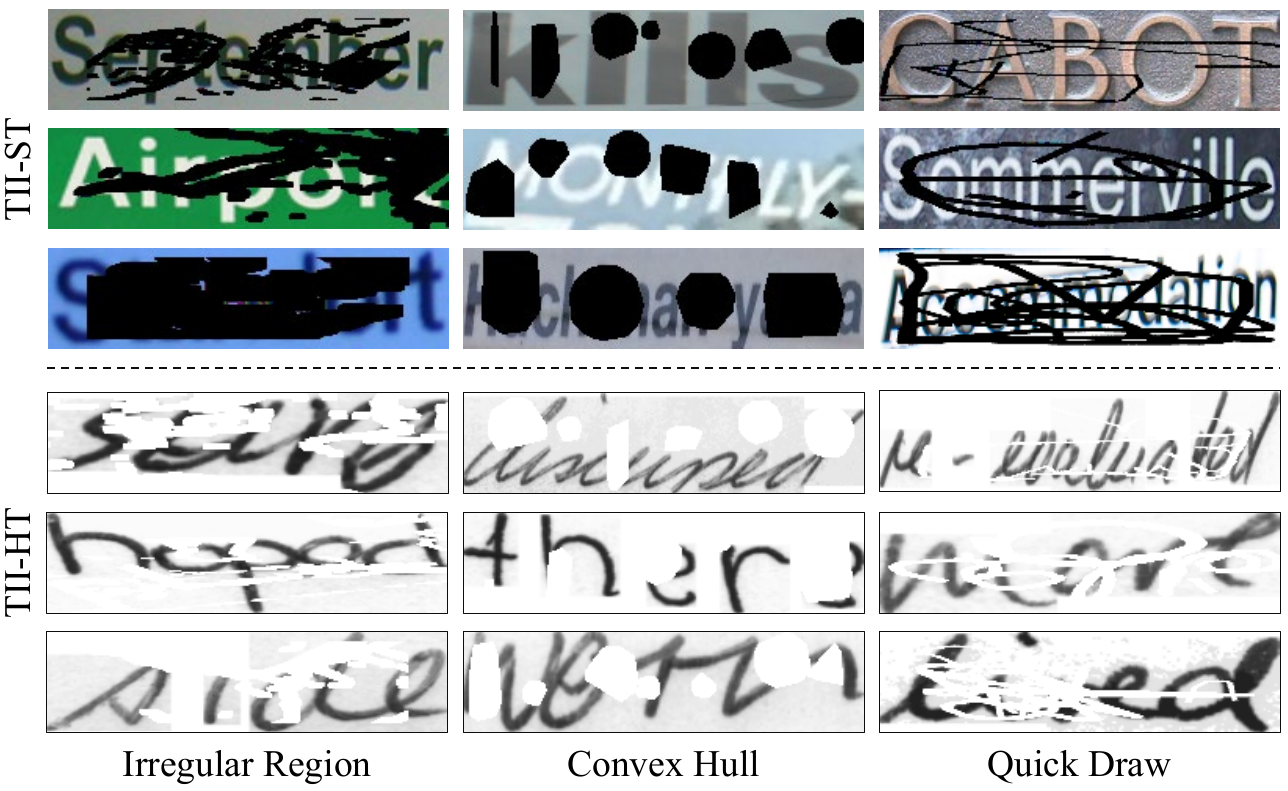} 
\caption{Illustration of samples in TII-ST and TII-HT.}
\label{fig:datasetfig}
\end{figure}

\begin{figure*}[ht]
\centering
\includegraphics[width=1\textwidth]{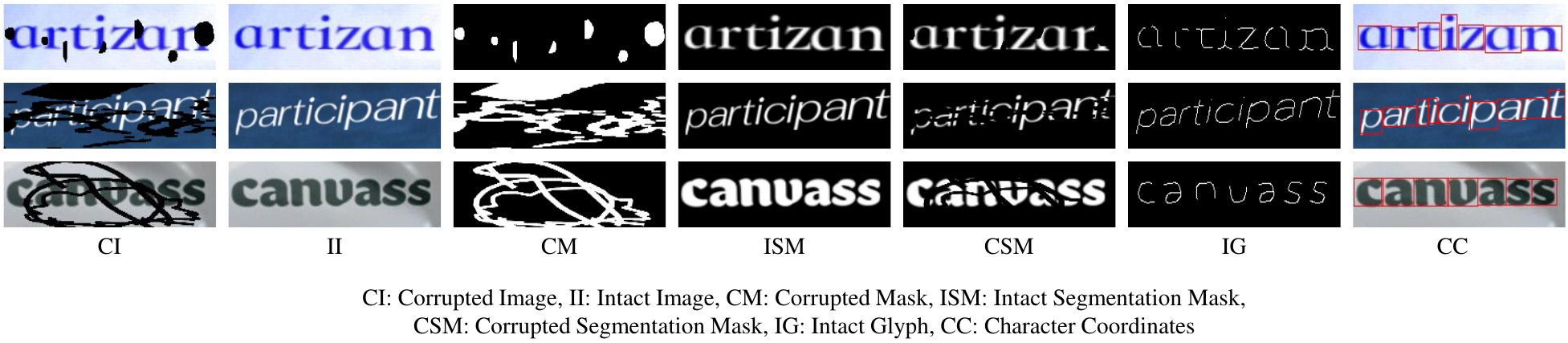} 
\caption{Illustration of all information for synthesis scene text images.}
\label{fig:assistant}
\end{figure*}

Here we introduce the construction of corrosion forms in detail:
(1) Convex Hull: This corrosion form is based on a geometric concept in which a shape is corrupted by an irregular convex polygon. We achieve this corrosion using the Graham algorithm~\cite{graham1972efficient}, mimicking the kind of damage that could occur from physical wear or removal of a portion of the image. (2) Irregular Region: Irregular Region: In this form, corrosion occurs within an irregularly shaped area. This can simulate more complex types of damage or corrosion that might occur in real-world scenarios, such as rust damage or non-uniform fading. We collect 12,000 masks from an existing dataset~\cite{liu2018image} and add data augmentation for diversity. (3) Quick Draw: Quick Draw: This form of corrosion mimics the effect of hastily drawn scribbles or marks, simulating the damage that might occur if someone scribbles or writes over the text image. We employ dilation operations on masks from the Quick Draw Irregular Mask Dataset (QD-IMD) to address complex situations. Notably, we designate the corrosion region as black for TII-ST and white for TII-HT to align with practical scenarios. Representative samples are displayed in Figure~\ref{fig:datasetfig}. Additionally, Figure~\ref{fig:assistant} provides supplementary information for the 80k synthesized scene text images, potentially aiding future text image inpainting developments.

\subsubsection{Detailed Statistics}
\begin{figure}[ht]
\centering
\includegraphics[width=1\columnwidth]{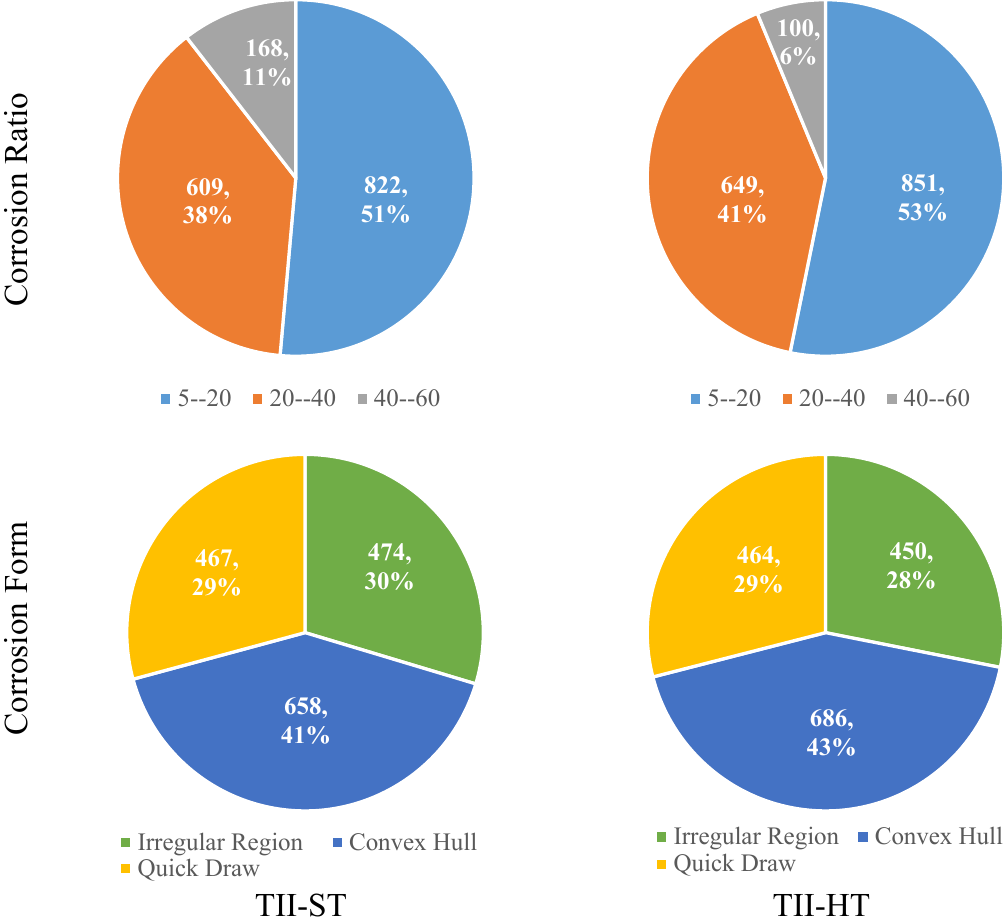} 
\caption{The statistics of corrosion ratio ranges and forms.}
\label{fig:Statistics}
\end{figure}

We provide a comprehensive overview of the corrosion ratio ranges and forms present in the testing sets of TII-ST and TII-HT.  As depicted in Figure~\ref{fig:Statistics}, the corrosion ratio spans a wide range, with each proportion being appropriate.

\section{Implementation Details}

\subsection{Architecture of GSDM}
In the Structure Prediction Module (SPM), we design a compact U-Net with three pairs of symmetric dilated convolution blocks. Each of these blocks incorporates two $3\times3$ dilated convolution layers (with a dilation rate of 2), a respective up/down-sampling layer, a Batch Normalization (BN) layer, and an ELU activation layer. 
Meanwhile, in the Reconstruction Module (RM), we adopt the classic U-Net structure, featuring five symmetric pairs of convolution blocks as detailed in~\cite{ho2020denoising}. Each block in this structure comprises one $3\times3$ standard convolution layer and two residual sub-blocks. Within each sub-block, there are one linear layer and two convolution modules, and each module houses a $3\times3$ standard convolution layer, a Group Normalization (GN) layer, a Swish activation layer, and a dropout layer.
Variations in image feature dimensions before/after processing through each block are detailed in Table~\ref{tab:gsdm}.

\begin{table}[ht]
\centering
\begin{tabular}{cll}
\Xhline{1.2pt}
Block & Input size & Output size \\ \hline
\multicolumn{3}{c}{SPM} \\ \hline
1 & $3 \times 64 \times 256$ &  $32 \times 64 \times 256$ \\
2 & $32 \times 64 \times 256$ & $64 \times 32 \times 128$ \\
3 & $64 \times 32 \times 128$ & $128 \times 16 \times 64$ \\
4 & $128 \times 16 \times 64$  &  $256 \times 8 \times 32$ \\
5 &  $256 \times 8 \times 32$ &  $128 \times 16 \times 64$ \\
6 &  $128 \times 16 \times 64$ & $64 \times 32 \times 128$ \\
7 &  $64 \times 32 \times 128$ & $32 \times 64 \times 256$ \\
8 &  $32 \times 64 \times 256$ & $3 \times 64 \times 256$ \\
\hline
\multicolumn{3}{c}{RM} \\ \hline
1 & $9 \times 64 \times 256$ & $64 \times 64 \times 256$\\
2 &$64 \times 64 \times 256$  & $128 \times 32 \times 128$ \\
3 &$128 \times 32 \times 128$ & $256 \times 16 \times 64$ \\
4 &$256 \times 16 \times 64$&$512 \times 8 \times 32$  \\
5 &$512 \times 8 \times 32$&$512 \times 4 \times 16$  \\
6 & $512 \times 4 \times 16$ &$512 \times 8 \times 32$\\
7 &$512 \times 8 \times 32$  & $256 \times 16 \times 64$ \\
8 &$256 \times 16 \times 64$& $128 \times 32 \times 128$ \\
9 &$128 \times 32 \times 128$  & $64 \times 64 \times 256$ \\
10 & $64 \times 64 \times 256$ &$3 \times 64 \times 256$  \\ \Xhline{1.2pt}
\end{tabular}
  \caption{Changes in image feature size after each block.}
  \label{tab:gsdm}%
\end{table}

\subsection{Training Details}

We implement all the comparison methods on the NVIDIA TITAN RTX 24G GPU using CUDA version 11.3. For consistency across evaluations, all input images, specifically the corrupted ones, for these methods are standardized to a size of $64 \times 256$. Accordingly, we make essential modifications to each method to adapt the requisite feature size.

For the open-source methods, namely CoPaint~\cite{zhang2023towards}, GDP~\cite{fei2023generative}, VCNet~\cite{wang2020vcnet}, and TransCNN-HAE~\cite{zhao2022transcnn}, we employ the default training setting and the official implementation of each one. For the methods that are not open-source, i.e., TSINIT~\cite{sun2022tsinit} and \cite{wang2021chinese}, we follow their methodologies and training settings for each model to reproduce the code. Notably, since \cite{wang2021chinese} aims to address the Chinese character inpainting task, we replace the Chinese-based BERT~\cite{devlin2018bert} with the English-based counterpart. Furthermore, we utilize the conditional version of vanilla DDIM~\cite{song2020denoising}, where the predicted target is noise, to demonstrate the improvement of our GSDM. For a fair comparison, we try to fine-tune its hyperparameters to achieve optimal performance (The time step is 1000 in training and the sampling step is 50 in inference).

For the proposed GSDM, we adopt slightly different training strategies for the two datasets. For TII-ST, both the SPM and RM are optimized independently. In the training phase, we first train the SPM on our 80k synthesized images for 50 epochs using the Adam optimizer~\cite{kingma2014adam}, setting the learning rate to $1\times10^{-4}$ and the mini-batch size at 32. The weight $\lambda$ of each loss for optimizing SPM is consistently set to 1. After training the SPM, we train the RM on both synthesized and real images for 400 epochs using the same optimizer. Here, the learning rate is set to $1\times10^{-3}$ with a mini-batch size of 2. The time step $T$ is set to 2000 during the training phase and the setting of $\alpha$ follows~\cite{ho2020denoising}. For TII-HT, the settings are identical except for the training dataset. In inference, the time step $\tau$ is set to 1 to enhance efficiency. 

\section{Extensive Comparison Experiments}
\begin{figure}[ht]
\centering
\includegraphics[width=1\columnwidth]{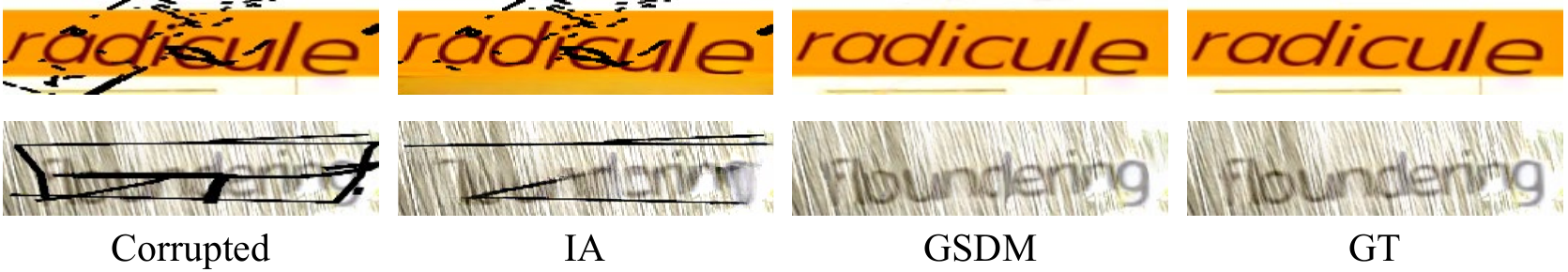} 
\caption{Illustration of the inpainting results on TII-ST. The ``IA" denotes the ``Inpainting Anything''.}
\label{fig:inpaintAnything}
\end{figure}

In our analysis, we extend our comparison of inpainting techniques on the TII-ST and TII-HT datasets. Specifically, we introduce GDP~\cite{fei2023generative}, a state-of-the-art method for blind image restoration, and VCNet~\cite{wang2020vcnet}, a representative approach for blind image inpainting. Furthermore, we assess the effectiveness of the large model-based method. As shown in Figure~\ref{fig:inpaintAnything}, the  ``Inpaint Anything''~\cite{yu2023inpaint} model, combined with the SAM~\cite{kirillov2023segment} and Stable Diffusion~\cite{rombach2022high}, falls short in addressing the challenges presented by our proposed task.

\subsection{Performance of Different Granularities}
Armed with these comparative methods, we introduce an additional metric: character-level recognition accuracy (abbr. Char Acc), modified from Character Error Rate. This metric can measure the improvement in recognition performance with a focus on finer granularity. For the $i$-th text image in the dataset $\mathcal{D}$, given the predicted text sequence $P_{i}$ and the ground truth text label $G_{i}$, the total Char Acc is computed as:
\begin{equation}
\mathrm{Char \ Acc}=\frac{1}{|\mathcal{D}|}\sum^{|\mathcal{D}|}_{i=1} (1- \frac{\mathrm{ED}\left( P_{i},G_{i} \right)}{\mathrm{Max}\left( |P_{i}|,|G_{i}| \right)}),
\label{eq:characc}
\end{equation}
where $\mathrm{ED(\cdot)}$ stands for the edit distance~\cite{levenshtein1966binary}, $|P_{i}|$, $|G_{i}|$ and $|\mathcal{D}|$ refer to the length of the prediction sequence, the length of the ground truth text label and the image number in the dataset $D$, respectively. A larger Char Acc implies that the predicted sequences closely match the ground-truth labels. Meanwhile, the word-level recognition accuracy (abbr. Word Acc) can be computed as:
\begin{equation}
\mathrm{Word \ Acc}=\frac{1}{{|\mathcal{D}|}}{\sum^{|\mathcal{D}|}_{i=1} \mathbb{I}(P_{i}=G_{i})},
\label{eq:wordacc}
\end{equation}
where $\mathbb{I}$ denotes the indicator function.

Based on the two metrics, we conduct two comparison experiments shown in Table~\ref{tab:comparision_ST_1} and Table~\ref{tab:comparision_HT_1}.
The results highlight that our proposed GSDM still delivers superior performance enhancements for downstream recognition tasks among inpainting techniques, both at coarse and fine granularities. Notably, VCNet cannot address handwritten text images and only generates all-zero matrices.

\begin{table}[ht]
\centering
\resizebox{1\linewidth}{!}{
\begin{tabular}{ll|cccc}
\Xhline{1.2pt}
Type & Method & Accuracy & PSNR & SSIM & Time (s) \\ \hline
\multirow{2}{*}{(i)} & TransCNN-HAE & 67.19 & 28.36 & 0.9164 & \textbf{0.017} \\
 & VCNet & 61.50 & 24.29 & 0.8709 & 0.074 \\ \hline
\multirow{4}{*}{(ii)} & DDIM & 55.35 & 16.79 & 0.7007 & 0.836 \\
 & CoPaint & 62.96 & 26.21 & 0.8794 & 72.850 \\ 
 & GDP & 47.05 & 21.46 & 0.7832 & 108.709 \\
 & GSDM (ours) & \textbf{71.73} & \textbf{33.28} & \textbf{0.9596} & 0.035 \\ \Xhline{1.2pt}
\end{tabular}}
\caption{The efficiency comparison of different methods on TII-ST. The ``Time'' represents the inference time cost of each method per image. Type (i) denotes the encoder-decoder-based model and type (ii) denotes the diffusion-based model. }
\label{tab:Effecency}
\end{table}

\begin{table*}[ht]
\centering
\begin{tabular}{l|cccccccc}
\Xhline{1.2pt}
Dataset & \multicolumn{8}{c}{TII-ST} \\ \hline
\multirow{2}{*}{Metric} & \multicolumn{2}{c}{CRNN} & \multicolumn{2}{c}{ASTER} & \multicolumn{2}{c|}{MORAN} & \multicolumn{2}{c}{Quality} \\
 & Word Acc & Char Acc & Word Acc & Char Acc & Word Acc & \multicolumn{1}{c|}{Char Acc} & PSNR & SSIM \\ \hline
Corrupted Image & 16.89 & 53.96 & 26.21 & 54.23 & 27.08 & \multicolumn{1}{c|}{59.27} & 14.24 & 0.7018 \\ \hline
TSINIT\dag & 56.54 & 82.91 & 63.60 & 85.85 & 61.22 & \multicolumn{1}{c|}{85.12} & - & - \\
DDIM & 50.59 & 79.03 & 60.73 & 81.89 & 58.53 & \multicolumn{1}{c|}{81.51} & 16.79 & 0.7007 \\
CoPaint* & 56.91 & 82.34 & 66.23 & 86.27 & 65.73 & \multicolumn{1}{c|}{86.23} & 26.21 & 0.8794 \\
TransCNN-HAE & 60.41 & 86.28 & 70.61 & 89.38 & 70.55 & \multicolumn{1}{c|}{89.70} & 28.36 & 0.9164 \\
VCNet & 54.28 & 83.39 & 65.36 & 87.15 & 64.86 & \multicolumn{1}{c|}{86.97} & 24.29 & 0.8709 \\
GDP & 39.28 & 73.47 & 51.47 & 76.69 & 50.41 & \multicolumn{1}{c|}{77.48} & 21.46 & 0.7832 \\
GSDM (ours) & \textbf{67.48} & \textbf{89.90} & \textbf{74.67} & \textbf{92.19} & \textbf{73.04} & \multicolumn{1}{c|}{\textbf{91.46}} & \textbf{33.28} & \textbf{0.9596} \\ \hline
Ground Truth & 80.18 & 94.10 & 88.74 & 96.69 & 86.93 & \multicolumn{1}{c|}{95.96} & - & - \\ \Xhline{1.2pt}
\end{tabular}
\caption{The comparison results on TII-ST, respectively. The ``-" denotes unavailable. ``*" and ``\dag" denote the non-blind method and reproduction version by ourselves, respectively.}
  \label{tab:comparision_ST_1}
\end{table*}

\begin{table*}[ht]
\centering
\begin{tabular}{l|cccccccc}
\Xhline{1.2pt}
Dataset & \multicolumn{8}{c}{TII-HT} \\ \hline
\multirow{2}{*}{Metric} & \multicolumn{2}{c}{DAN} & \multicolumn{2}{c}{TrOCR-L} & \multicolumn{2}{c|}{TrOCR-B} & \multicolumn{2}{c}{Quality} \\
 & Word Acc & Char Acc & Word Acc & Char Acc & Word Acc & \multicolumn{1}{c|}{Char Acc} & PSNR & SSIM \\ \hline
Corrupted Image & 23.81 & 60.53 & 33.25 & 61.91 & 19.75 & \multicolumn{1}{c|}{50.71} & 20.09 & 0.8916 \\ \hline
Wang et al.\dag & 21.63 & 57.57 & 18.50 & 47.57 & 11.00 & \multicolumn{1}{c|}{37.85} & 16.89 & 0.8113 \\
DDIM & 0.25 & 0.66 & 44.13 & 71.37 & 10.75 & \multicolumn{1}{c|}{40.48} & 9.32 & 0.2842 \\
CoPaint* & 42.12 & 74.23 & 45.50 & 71.60 & 26.06 & \multicolumn{1}{c|}{56.00} & 24.52 & 0.9203  \\
TransCNN-HAE & 17.19 & 58.22 & 47.25 & 73.71 & 22.87 & \multicolumn{1}{c|}{53.19} & 15.42 & 0.7675 \\
GDP & 16.82 & 49.89 & 33.63 & 61.11 & 14.50 & \multicolumn{1}{c|}{45.17} & 17.61 & 0.7506 \\
GSDM (ours) & \textbf{69.43} & \textbf{89.85} & \textbf{66.81} & \textbf{86.09} & \textbf{56.00} & \multicolumn{1}{c|}{\textbf{79.72}} & \textbf{32.13} & \textbf{0.9718} \\ \hline
Ground Truth & 85.19 & 95.37 & 75.56 & 90.94 & 64.07 & \multicolumn{1}{c|}{84.86} & - & - \\ \Xhline{1.2pt}
\end{tabular}
\caption{The comparison results on TII-HT, respectively. The ``-" denotes unavailable. ``*" and ``\dag" denote the non-blind method and reproduction version by ourselves, respectively.}
  \label{tab:comparision_HT_1}
\end{table*}

\subsection{Performance on Different Types of Data}

Since our datasets can be segmented by region ratio and form of corrosion, we conducted detailed comparative experiments to thoroughly assess our proposed method. Notably, the metric ``Accuracy" refers to the average word accuracy of three recognition models (CRNN, ASTER, MORAN for STR and DAN, TrOCR-B, TrOCR-L for HTR), consistent with the experiments presented in the main paper.

\subsubsection{Performance of Different Corrosion Ratios}

Table~\ref{tab:comparision_ST_ratio} and Table~\ref{tab:comparision_HT_ratio} depict the inpainting performance on images segmented by different corrosion ratios. From the results, we can infer the following:
(1) On subsets with minor corrosion areas (5\%--20\%), our method approaches the performance of real images. This suggests that, unlike other generative models, our method does not deteriorate the image quality. (2) As the ratio of the corrosion region increases, the inpainting efficacy of some comparative methods, such as VCNet on TII-ST and CoPaint on TII-HT, markedly diminishes. In stark contrast, our method consistently exhibits superior performance.

\begin{table*}[ht]
\centering
\begin{tabular}{l|ccccccccc}
\Xhline{1.2pt}
Dataset & \multicolumn{9}{c}{TII-ST} \\ \hline
Ratio & \multicolumn{3}{c|}{5\%--20\%} & \multicolumn{3}{c|}{20\%--40\%} & \multicolumn{3}{c}{40\%--60\%} \\ \hline
Metric & Accuracy & PSNR & \multicolumn{1}{c|}{SSIM} & Accuracy & PSNR & \multicolumn{1}{c|}{SSIM} & Accuracy & PSNR & SSIM \\ \hline
Corrupted Image & 39.13 & 16.74 & \multicolumn{1}{c|}{0.8177} & 8.05 & 12.11 & \multicolumn{1}{c|}{0.6183} & 1.98 & 9.74 & 0.4377 \\ \hline
TSINIT\dag & 73.52 & - & \multicolumn{1}{c|}{-} & 53.91 & - & \multicolumn{1}{c|}{-} & 20.24 & - & - \\
DDIM & 69.87 & 17.56 & \multicolumn{1}{c|}{0.7199} & 48.00 & 16.45 & \multicolumn{1}{c|}{0.6988} & 23.02 & 14.34 & 0.6137 \\
CoPaint* & 78.91 & 30.08 & \multicolumn{1}{c|}{0.9427} & 51.12 & 23.12 & \multicolumn{1}{c|}{0.8448} & 25.20 & 18.52 & 0.6956 \\
TransCNN-HAE & 80.01 & 32.13 & \multicolumn{1}{c|}{0.9584} & 59.11 & 25.29 & \multicolumn{1}{c|}{0.8924} & 33.73 & 21.09 & 0.7978 \\
VCNet & 77.74 & 27.35 & \multicolumn{1}{c|}{0.9262} & 50.36 & 23.00 & \multicolumn{1}{c|}{0.8399} & 22.42 & 17.82 & 0.7127 \\
GDP & 66.87 & 24.29 & \multicolumn{1}{c|}{0.8650} & 31.36 & 19.22 & \multicolumn{1}{c|}{0.7314} & 6.94 & 15.71 & 0.5706 \\
GSDM (ours) & \textbf{83.42} & \textbf{38.12} & \multicolumn{1}{c|}{\textbf{0.9883}} & \textbf{65.79} & \textbf{29.61} & \multicolumn{1}{c|}{\textbf{0.9503}} & \textbf{36.11} & \textbf{22.98} & \textbf{0.8531} \\ \hline
Ground Truth & 85.40 & - & \multicolumn{1}{c|}{-} & 84.46 & - & \multicolumn{1}{c|}{-} & 87.90 & - & - \\ \Xhline{1.2pt}
\end{tabular}
\caption{The comparison results of different corrosion region ratios on TII-ST. The ``-" denotes unavailable. ``*" and ``\dag" denote the non-blind method and reproduction version by ourselves, respectively.}
  \label{tab:comparision_ST_ratio}
\end{table*}

\begin{table*}[]
\centering
\begin{tabular}{l|ccccccccc}
\Xhline{1.2pt}
Dataset & \multicolumn{9}{c}{TII-HT} \\ \hline
Ratio & \multicolumn{3}{c|}{5\%--20\%} & \multicolumn{3}{c|}{20\%--40\%} & \multicolumn{3}{c}{40\%--60\%} \\ \hline
Metric & Accuracy & PSNR & \multicolumn{1}{c|}{SSIM} & Accuracy & PSNR & \multicolumn{1}{c|}{SSIM} & Accuracy & PSNR & SSIM \\ \hline
Corrupted Image & 40.03 & 22.47 & \multicolumn{1}{c|}{0.9363} & 10.43 & 17.70 & \multicolumn{1}{c|}{0.8529} & 1.33 & 15.22 & 0.7617 \\ \hline
Wang et al.\dag & 27.38 & 17.93 & \multicolumn{1}{c|}{0.8432} & 6.01 & 15.91 & \multicolumn{1}{c|}{0.7839} & 0.67 & 14.47 & 0.7180 \\
DDIM & 23.62 & 9.39 & \multicolumn{1}{c|}{0.2949} & 13.82 & 9.30 & \multicolumn{1}{c|}{0.2786} & 3.33 & 8.83 & 0.2289 \\
CoPaint* & 55.46 & 27.83 & \multicolumn{1}{c|}{0.9606} & 20.13 & 21.24 & \multicolumn{1}{c|}{0.8889} & 3.67 & 17.68 & 0.7814 \\
TransCNN-HAE & 34.90 & 15.57 & \multicolumn{1}{c|}{0.7805} & 24.70 & 15.29 & \multicolumn{1}{c|}{0.7574} & 8.33 & 15.01 & 0.7219 \\
GDP & 32.71 & 19.90 & \multicolumn{1}{c|}{0.8235} & 10.43 & 15.17 & \multicolumn{1}{c|}{0.6815} & 0.33 & 13.83 & 0.5792 \\
GSDM (ours) & \textbf{72.81} & \textbf{37.22} & \multicolumn{1}{c|}{\textbf{0.9923}} & \textbf{58.81} & \textbf{27.25} & \multicolumn{1}{c|}{\textbf{0.9594}} & \textbf{24.00} & \textbf{20.46} & \textbf{0.8775} \\ \hline
Ground Truth & 75.44 & - & \multicolumn{1}{c|}{-} & 74.73 & - & \multicolumn{1}{c|}{-} & 72.00 & - & - \\ \Xhline{1.2pt}
\end{tabular}
\caption{The comparison results of different corrosion region ratios on TII-HT. The ``-" denotes unavailable. ``*" and ``\dag" denote the non-blind method and reproduction version by ourselves.}
  \label{tab:comparision_HT_ratio}
\end{table*}

\subsection{Comparison of Efficiency}

In Table~\ref{tab:Effecency}, we compare the efficiency of various inpainting methods on TII-ST. Note that the time cost (0.034s) mentioned in the ablation experiment section of the main text accounts only for the processing time of the RM module. Thanks to the lightweight SPM model architecture, the overall time cost increases by only 0.001s. Our observations from the results are twofold: (1) Due to the one-step inference strategy within the RM, our proposed GSDM significantly outshines other diffusion-based models in efficiency. (2) Regarding time cost magnitude, our method stands on par with encoder-decoder-based methods. However, the inpainting performance of GSDM is markedly superior to them.

\subsubsection{Performance of Different Corrosion Forms}

Table~\ref{tab:comparision_ST_form} and Table~\ref{tab:comparision_HT_form} present the inpainting performance on images segmented by different corrosion forms. Results indicate that: (1) While the performance of our method on the convex hull form (TII-ST) is marginally surpassed by TransCNN-HAE, it excels in managing irregular regions and quick draw forms, significantly outperforming other comparative algorithms. (2) Our approach excels particularly in restoring the quick draw form, generating inpainting results that closely resemble real images, a challenge that other methods struggle to address.


\begin{table*}[ht]
\centering
\begin{tabular}{l|ccccccccc}
\Xhline{1.2pt}
Dataset & \multicolumn{9}{c}{TII-ST} \\ \hline
Form & \multicolumn{3}{c|}{Convex Hull} & \multicolumn{3}{c|}{Irregular Region} & \multicolumn{3}{c}{Quick Draw} \\ \hline
Metric & Accuracy & PSNR & \multicolumn{1}{c|}{SSIM} & Accuracy & PSNR & \multicolumn{1}{c|}{SSIM} & Accuracy & PSNR & SSIM \\ \hline
Corrupted Image & 32.42 & 14.97 & \multicolumn{1}{c|}{0.7802} & 17.72 & 13.35 & \multicolumn{1}{c|}{0.6526} & 16.42 & 14.11 & 0.6413 \\ \hline
TSINIT\dag & 56.99 & - & \multicolumn{1}{c|}{-} & 57.81 & - & \multicolumn{1}{c|}{-} & 68.02 & - & - \\
DDIM & 52.48 & 16.60 & \multicolumn{1}{c|}{0.7097} & 58.30 & 16.72 & \multicolumn{1}{c|}{0.6997} & 60.74 & 17.15 & 0.6891 \\
CoPaint* & 54.51 & 23.90 & \multicolumn{1}{c|}{0.8656} & 61.67 & 25.94 & \multicolumn{1}{c|}{0.8611} & 75.23 & 29.75 & 0.9177 \\
TransCNN-HAE & \textbf{66.62} & 31.09 & \multicolumn{1}{c|}{0.9509} & 63.64 & 25.59 & \multicolumn{1}{c|}{0.8814} & 71.59 & 27.34 & 0.9033 \\
VCNet & 56.33 & 24.07 & \multicolumn{1}{c|}{0.8831} & 60.97 & 23.55 & \multicolumn{1}{c|}{0.8513} & 69.31 & 25.34 & 0.8736 \\
GDP & 49.90 & 22.14 & \multicolumn{1}{c|}{0.8247} & 41.00 & 20.59 & \multicolumn{1}{c|}{0.7525} & 49.18 & 21.39 & 0.7558 \\
GSDM (ours) & 64.08 & \textbf{31.48} & \multicolumn{1}{c|}{\textbf{0.9511}} & \textbf{73.77} & \textbf{32.82} & \multicolumn{1}{c|}{\textbf{0.9554}} & \textbf{80.44} & \textbf{36.31} & \textbf{0.9760} \\ \hline
Ground Truth & 85.16 & - & \multicolumn{1}{c|}{-} & 86.85 & - & \multicolumn{1}{c|}{-} & 83.94 & - & - \\ \Xhline{1.2pt}
\end{tabular}
\caption{The comparison results of different corrosion forms on TII-ST. The ``-" denotes unavailable. ``*" and ``\dag" denote the non-blind method and reproduction version by ourselves, respectively.}
  \label{tab:comparision_ST_form}
\end{table*}

\begin{table*}[ht]
\centering
\begin{tabular}{l|ccccccccc}
\Xhline{1.2pt}
Dataset & \multicolumn{9}{c}{TII-HT} \\ \hline
Form & \multicolumn{3}{c|}{Convex Hull} & \multicolumn{3}{c|}{Irregular Region} & \multicolumn{3}{c}{Quick Draw} \\ \hline
Metric & Accuracy & PSNR & \multicolumn{1}{c|}{SSIM} & Accuracy & PSNR & \multicolumn{1}{c|}{SSIM} & Accuracy & PSNR & SSIM \\ \hline
Corrupted Image & 26.29 & 20.57 & \multicolumn{1}{c|}{0.9048} & 22.00 & 19.56 & \multicolumn{1}{c|}{0.8770} & 28.09 & 19.87 & 0.8860 \\ \hline
Wang et al.\dag & 16.09 & 17.05 & \multicolumn{1}{c|}{0.8154} & 16.44 & 16.71 & \multicolumn{1}{c|}{0.8034} & 19.04 & 16.84 & 0.8130 \\
DDIM & 17.25 & 9.26 & \multicolumn{1}{c|}{0.2810} & 18.52 & 9.31 & \multicolumn{1}{c|}{0.2829} & 19.90 & 9.40 & 0.2900 \\
CoPaint* & 33.72 & 23.84 & \multicolumn{1}{c|}{0.9139} & 34.96 & 23.95 & \multicolumn{1}{c|}{0.9105} & 46.91 & 26.08 & 0.9393 \\
TransCNN-HAE & 26.14 & 15.41 & \multicolumn{1}{c|}{0.7644} & 28.97 & 15.42 & \multicolumn{1}{c|}{0.7653} & 33.62 & 15.44 & 0.7741 \\
GDP & 22.93 & 18.22 & \multicolumn{1}{c|}{0.7961} & 19.63 & 16.55 & \multicolumn{1}{c|}{0.7129} & 21.70 & 17.72 & 0.7199 \\
GSDM (ours) & \textbf{58.99} & \textbf{29.93} & \multicolumn{1}{c|}{\textbf{0.9620}} & \textbf{62.52} & \textbf{31.80} & \multicolumn{1}{c|}{\textbf{0.9705}} & \textbf{73.13} & \textbf{35.69} & \textbf{0.9875} \\ \hline
Ground Truth & 75.56 & - & \multicolumn{1}{c|}{-} & 74.07 & - & \multicolumn{1}{c|}{-} & 74.86 & - & - \\ \Xhline{1.2pt}
\end{tabular}
\caption{The comparison results of different corrosion forms on TII-HT. The ``-" denotes unavailable. ``*" and ``\dag" denote the non-blind method and reproduction version by ourselves, respectively.}
  \label{tab:comparision_HT_form}
\end{table*}

\end{document}